\documentclass[journal,11pt,onecolumn]{IEEEtran}
\usepackage[margin=1.25in]{geometry}
\usepackage{graphicx}
\usepackage{header}
\usepackage{url}
\usepackage{subfig}
\definecolor{sorange}{rgb}{0.971, 0.57, 0.025}

\newif\ifsc
\sctrue

\begin{document}
%
\title{Inverse Risk-Sensitive Reinforcement Learning}
%
%
%

\author{Lillian J. Ratliff and Eric Mazumdar
     \thanks{L.~Ratliff  is  with the Department
of Electrical Engineering, University of Washington, Seattle,
WA, 98185. email: {\tt ratliffl@uw.edu}.}
\thanks{E.~Mazumdar is with the Department of Electrical Engineering and
Computer Sciences at the University of California, Berkeley, Berkeley, CA 94720.
email: {\tt mazumdar@berkeley.edu}}}
\maketitle

\begin{abstract}
    We address the problem of inverse reinforcement learning in
Markov decision processes where the agent is risk-sensitive. 
We derive a risk-sensitive reinforcement learning algorithm with convergence
guarantees that 
employs convex risk metrics and models of human decision-making
deriving from behavioral economics.
The risk-sensitive reinforcement learning algorithm 
provides the  
theoretical underpinning
for
a gradient-based inverse reinforcement
learning algorithm that minimizes a loss function defined on observed behavior
of a risk-sensitive agent.
We demonstrate the performance of the proposed technique on two examples: (i)
the canonical Grid World example and (ii) 
a
Markov decision process modeling ride-sharing passengers' decisions given price changes. In the latter, we use
pricing and travel time data from a ride-sharing company to construct the transition
probabilities and rewards of the Markov decision process.

\end{abstract}


%
\IEEEpeerreviewmaketitle

\section{Introduction}
\label{sec:intro}
\IEEEPARstart{T}{he} modeling  and learning of human decision-making
behavior is increasingly becoming important as critical systems begin to rely more on automation and artificial intelligence.
Yet,
in this task we face a number of challenges, not least of which is the fact
that 
humans are known to behave in ways that are not completely rational. There
is mounting evidence to support the fact that humans often use \emph{reference
points}---e.g., the \emph{status quo} or former experiences or recent
expectaions about the future that are otherwise perceived to be related to the decision the human is
making~\cite{koszegi:2006aa,tversky:1991aa}. It has also been observed that their decisions are impacted by their perception of the
external world (exogenous factors) and their present state of mind (endogenous
factors) as well as how the decision is \emph{framed} or
presented~\cite{tversky:1986aa}.

The success of \emph{descriptive} behavioral models in capturing human behavior has long been touted by the psychology
community and, more recently, by the
economics community.  In the engineering context, humans have largely been
modeled, under rationality assumptions, from the so-called \emph{normative} point of
view where things are modeled \emph{as they ought to be}, which is counter to a
descriptive \emph{as is} point of view.

However, risk-sensitivity in the context of learning to control
stochastic dynamical
systems (see,
\emph{e.g.},~\cite{geibel:2005aa, borkar:2002ab}) has
been fairly extensively explored in engineering. Many of these approaches are targeted at mitigating risks due
to uncertainties in controlling a system such as a plant or
robot
where \emph{risk-aversion} is captured by leveraging techniques such as exponential utility
functions or minimizing mean-variance criteria. 

Complex risk-sensitive behavior arising from
human interaction with automation is only recently coming into focus. Human decision makers can
be at once risk-averse and risk-seeking depending their frame of reference.
 The
adoption of diverse behavioral models in engineering---in particular, in learning
and control---is growing due to the fact that humans are increasingly
playing an integral role in automation both at the individual and societal
scale.
Learning accurate models of human decision-making is important for both
\emph{prediction} and \emph{description}. For example, control/incentive schemes
need to predict human behavior as a function of external stimuli including not
only potential disturbances but also the control/incentive mechanism itself. On the other
hand,
policy makers and regulatory agencies, {e.g.}, are interested in interpreting human reactions to
implemented regulations and policies.

Approaches for integrating the risk-sensitivity in the
control and reinforcement
learning problems via behavioral
models have
recently emerged~\cite{l.a.:2016aa,shen:2014aa,mihatsch:2002aa,nagengast:2010aa, majumdar:2017aa}. These
approaches largely assume a risk-sensitive Markov decision process (MDP) formulated based on
a model that captures behavioral aspects of the human's decision-making process.
We refer the problem of learning the optimal policy in this setting as
the \emph{forward} problem. 
Our primary interest
is 
in solving the
so-called \emph{inverse} problem which seeks to estimate the decision-making
process given a set of demonstrations; yet, to do so requires a well formulated
forward problem with convergence guarantees.

Inverse reinforcement learning in the
context of recovering policies directly (or indirectly via first learning a
representation for the reward) has long been studied in the context expected
utility maximization and MDPs~\cite{ng:2000aa,abbeel:2004aa,ratliff:2006aa}.  We may care about, e.g., producing
the value and reward
        functions (or at least, characterize the space of these functions) 
        that
        produce behaviors matching that which is observed. On the other hand, we may want to extract the
        optimal policy from a set of demonstrations so that we can reproduce the
        behavior in support of, {e.g.}, designing incentives or control
        policies. In this paper, our focus is on the combination of these two tasks.

       We model human decision-makers as
\emph{risk-sensitive Q-learning agents} where
we exploit very rich behavioral models from behavioral psychology and economics
that capture a whole spectrum of
risk-sensitive behaviors and loss aversion. We first derive a reinforcement
learning algorithm that leverages convex risk metrics and behavioral value
functions. We provide
convergence guarantees via a contraction mapping argument. In comparison to
previous work in this area~\cite{shen:2013aa}, we show that the behavioral value functions we introduce satisfy
the assumptions of our theorems.  

Given the forward risk-sensitive reinforcement learning algorithm, we propose a gradient-based learning algorithm for inferring
the decision-making model parameters from demonstrations---that is, we propose a framework for solving the
\emph{inverse risk-sensitive  reinforcement learning} problem with theoretical
guarantees. We show that the
gradient of the loss function with respect to the model parameters is
well-defined and computable
via a contraction map argument. We demonstrate the
efficacy of the learning scheme on the canonical Grid World example and a
passenger's view of ride-sharing modeled as an MDP with parameters estimated from
real-world
data.

 The work in this paper significantly extends our previous
        work~\cite{mazumdar:2017aa} first, by providing the proofs for the theoretical
        results appearing in the earlier work and second, by providing a more
        extensive theory for both the forward and inverse risk-sensitive reinforcement
        problems.

The remainder of this paper is organized as follows. In
 Section~\ref{sec:rsrl}, we overview the model we assume for
 risk-sensitive agents, show that it is amenable to integration with the
 behavioral models, and present our risk-sensitive Q-learning convergence results.   
 In Section~\ref{sec:rsirl}, we
 formulate the {\IRL} problem and propose a gradient--based algorithm to solve it. 
 Examples that demonstrate the ability of the proposed scheme to capture a
 wide breadth of risk-sensitive behaviors are provided in
 Section~\ref{sec:examples}. We comment on connections to recent related work in
 Section~\ref{sec:relatedwork}.
 Finally, we conclude with some discussion in
 Section~\ref{sec:discussion}.



\section{Risk-Sensitive Reinforcement Learning}
\label{sec:rsrl}
In order to learn a decision-making model for an agent who faces sequential
decisions in an uncertain environment, we leverage a risk-sensitive Q-learning
model that integrates coherent risk metrics with behavioral models. In
particular, the model we
use is based on a model first introduced in~\cite{heger:1994aa} and later
refined in~\cite{mihatsch:2002aa,shen:2014aa}.


The
primary difference between the work presented in this section and previous
work\footnote{For further
details on the relationship the work in this paper and related works, including
our previous work, see Section~\ref{sec:relatedwork}.
}
is that we (i) introduce a new prospect theory based value
function and (ii) provide a convergence theorem whose assumptions are
satisfied for the behavioral models we use. Under the assumption that the agent is making decisions
according to this model, in the sequel we formulate a gradient--based method for learning
the policy as well as parameters of the agent's value function. 

\subsection{Markov Decision Process}
We consider a class of finite MDPs consisting of
a
state space $X$, an admissible action space $A(x)\subset A$ for each $x\in X$, a
transition kernel $P(x'|x,a)$ that denotes the probability of moving from
state $x$ to $x'$ given action $a$, and a reward function\footnote{We note that
    it is possible to consider the more general reward structure $r:X\times
    A\times X\times W\rar\mb{R}$, however we exclude this case in order to not further bog down the notation.} $r:X\times A\times W\rar
\mb{R}$ where $W$ is the space of bounded disturbances and has
distribution $P_r(\cdot|x,a)$. Including disturbances allows us to model random
rewards; we use the notation $R(x',a)$ to denote the random
reward having distribution $P_r(\cdot|x,a)$.

In the classical expected utility maximization framework, the agent seeks to
maximize the expected discounted rewards by selecting a Markov policy $\pi$---that is, for an infinite horizon MDP, the optimal policy
 is obtained by maximizing
 \begin{equation}
     \textstyle J(x_0)=\max_{\pi}\mb{E}\left[ \sum_{t=1}^\infty\gamma^t
     R(x_{t},a_t)\right]
     \label{eq:exputil}
 \end{equation}
 where $x_0$ is the initial state and $\gamma\in(0,1)$ is the discount factor.
  
     The risk-sensitive {\RL} problem transforms the above problem to account for a
 salient features of the human decision-making process such as loss
aversion, reference point dependence, and risk-sensitivity.
Specifically, we introduce two key components, \emph{value functions} and
\emph{valuation functions}, that allow for our model to
 capture these features. The former captures
 risk-sensitivity, loss-aversion, and reference point dependence in its transformation of outcome values to
 their value as 
 perceived by the agent and the latter generalizes the expectation operator to more
 general measures of risk---specifically, \emph{convex risk measures}.

\subsection{Value Functions}
\label{sec:value}
     Given the environmental and reward uncertainties, we model
     the outcome of each action as a real-valued random variable $Y(i)\in \mb{R}$, $i\in
     I$ where $I$
denotes a finite event space and $Y$ is the outcome of $i$--th
event with probability $\mu(i)$ where $\mu\in\Delta(I)$, the space of probability
distributions on $I$.
Analogous to the expected utility framework, agents make choices based
on the value of the outcome determined by a \emph{value function}
$\uv:\mb{R}\rar\mb{R}$.

\label{sec:dm}
There are a number of existing approaches to defining value functions that capture risk-sensitivity and loss aversion. 
These approaches derive from a variety of fields including behavioral
psychology/economics, mathematical finance, and even neuroscience. 

One of the principal features of human decision-making is that losses are perceived more
significant than a gain of equal true value. The models with the greatest
efficacy in capturing this effect are
convex and concave in different regions of the outcome space. Prospect theory,
e.g., is built on one such model~\cite{kahneman:1979aa,tversky:1992aa}. The 
value function most commonly used in prospect theory is given by
\begin{equation}
    \uv(y)=\left\{\begin{array}{ll} \ \ \ku(y-\rp)^{\Lu}, & \ \  y>\rp\\
        -\kl(\rp-y)^{\Ll}, & \ \  y\leq
        \rp\end{array}\right.
    \label{eq:prospectu}
\end{equation}
where $\rp$ is the \emph{reference point} that the decision-maker compares outcomes
against in determining if the decision is a loss or gain.
The parameters $(\ku,\kl, \Lu,\Ll)$ control the degree of loss-aversion and
risk-sensitivity; e.g.,
\begin{enumerate}
    \item[(i)] $0<\Lu,\Ll<1$ implies preferences that are  risk-averse on gains and risk-seeking on losses
(concave in gains, convex in losses);
\item[(ii)]  $\Lu=\Ll=1$ implies
risk-neutral preferences;
\item[(iii)] $\Lu,\Ll>1$ implies preferences that are
risk-averse
on losses and risk-seeking on gains
(convex in gains, concave in losses).
\end{enumerate} 
%
%

Experimental results for a series of one-off decisions have indicated that
typically 
$0<\Lu, \Ll<1$ 
thereby indicating that humans are
risk-averse on gains and risk-seeking on losses.
%

In addition to the non-linear transformation of outcome values, in prospect
theory the effect of under/over-weighting the likelihood of events that has been
commonly observed in human behavior is modeled via \emph{warping} of event
probabilities~\cite{gonzalez:1999aa}. Other  concepts such as framing, reference dependence, and
loss aversion---captured, \emph{e.g.}, in the $(\ku,\kl)$ parameters in
\eqref{eq:prospectu}---have also been
widely observed in experimental studies 
(see,
\emph{e.g.},~\cite{simon:2000aa,tversky:1981aa,camerer:1989aa}).


Outside of the prospect theory value function, other mappings have been proposed to
capture risk-sensitivity. 
Proposed
in~\cite{mihatsch:2002aa}, 
the linear mapping 
\begin{equation}
    \uv(y)=\left\{\begin{array}{ll} (1-\kappa)y, & y>\rp\\ (1+\kappa)y, & y\leq
        \rp\end{array}\right.
    \label{eq:linearu}
\end{equation}
with
$\kappa\in(-1,1)$ is one such example. 
This value function can be viewed as a special case of \eqref{eq:prospectu}.

Another example is the entropic map which is given by
\begin{equation}
    \uv(y)=\exp(\lambda y)
    \label{eq:entropicmapvalue}
\end{equation}
 where $\lambda$
controls the degree of risk-sensitivity. The entropic map, however, is either
convex or concave on the entire outcome space.


%



Motivated by the empirical evidence supporting the prospect theoretic value
function and numerical considerations of our algorithm,
we introduce a value function that 
retains the shape of the prospect theory value function
while improving
the performance (in terms of convergence speed) of the forward and inverse
reinforcement learning procedures we propose.
In particular, we define the locally Lipschitz-prospect
(\lprospect) value function given by
\begin{equation}
    \uv(y)=\left\{\begin{array}{ll}\ \
        \ku(y-\rp+\epsilon)^{\Lu}-\ku\epsilon^{\Lu}, & \  y>\rp\\
        -\kl(\rp-y+\epsilon)^{\Ll}+\kl\epsilon^{\Ll},  & \  y\leq
        \rp\end{array}\right.
    \label{eq:exp}
\end{equation}
with $\ku,\kl, \Lu,\Ll>0$ and $\epsilon>0$, a small constant.
This value function is Lipschitz continuous on a bounded domain.
Moreover, the derivative of the {\lprospect} function is bounded away from zero
at the reference point. 
Hence, in practice it has better numerical properties.

We remark that, for given parameters $(\ku,\kl,\Lu,\Ll)$, the {\lprospect}
function has the same risk-sensitivity as the prospect value function with those
same parameters. Moreover, as $\epsilon\rar 0$ the {\lprospect} value function
approaches the prospect value function and thus, qualitatively speaking, the degree of Lipschitzness
decreases as $\epsilon\rar0$.

 The fact that each of these value
functions are defined by a small number of parameters that are highly
interpretable in terms of risk-sensitivity and loss-aversion is one of the
motivating factors for integrating them into a 
reinforcement learning
framework. It is our aim to
design learning algorithms that will ultimately provide the theoretical
underpinnings 
for designing incentives and control policies taking into consideration
salient features of human decision-making behavior.


\subsection{Valuation Functions via Convex Risk Metrics}
To further capture risk-sensitivity, \emph{valuation functions} generalize the
expectation operator, which 
 considers \emph{average} or \emph{expected} outcomes,\footnote{In the case of
     two events, the valuation function can also capture warping of
     probabilities. Alternative approaches to {\RL} based on cumulative prospect
     theory  for the more general case have been examined~\cite{l.a.:2016aa}.}
     to measures of risk.
     
     \begin{defn}[Monetary Risk Measure~\cite{follmer:2002aa}]
         A functional $\rho:\mc{X}\rar\mb{R}\cup\{+\infty\}$ on the space
         $\mc{X}$ of measurable functions defined on a probability space
         $(\Omega,\mc{F}, P)$ is said to be a monetary risk measure if $\rho(0)$
         is finite and if, for all $X,X'\in \mc{X}$, $\rho$
         satisfies the following: 
         \begin{enumerate} 
             \item (monotone) $X\leq X'$ $\Longrightarrow$ $\rho(X)\leq \rho(X')$
                          \item (translation invariant) $m\in \mb{R}$ $\Longrightarrow$
                 $\rho(X+m)=\rho(X)+m$
         \end{enumerate}
             \end{defn}
             If a monetary risk measure $\rho$ satisfies
         \begin{equation}
             \rho(\lambda X+(1-\lambda)X')\leq \lambda
             \rho(X)+(1-\lambda)\rho(X'),
             \label{eq:convexrequirement}
         \end{equation}
         for $\lambda\in [0,1]$, then it is a \emph{convex risk measure}. If,
         additionally, $\rho$ 
    is \emph{positive homogeneous}, i.e.~if $\lambda\geq 0$, then $\rho(\lambda
    X)=\lambda\rho(X)$, 
         then we call $\rho$ a \emph{coherent risk measure}.    
         While the results apply to coherent risk measures, we will primarily
         focus on convex measures of risk, a less restrictive class, that are generated by a set of
         \emph{acceptable} positions.
         
         Denote the space of probability measures on $(\Omega, \mc{F})$ by
         $\mc{M}_1(\Omega, \mc{F})$.
         \begin{defn}[Acceptable Positions]
            Consider a value function $\uv$, a probability measure $P\in
            \mc{M}_1(\Omega, \mc{F})$, and an \emph{acceptance level} $\uv_0=\uv(\bar{y})$ with
            $\bar{y}$ in the domain of $\uv$. The set 
    \begin{equation}
        \mc{A}=\{X\in \mc{X}|\ \mb{E}_P[\uv(X)]\geq \uv_0\}.
        \label{eq:acceptablepos}
    \end{equation}
    is the set of \emph{acceptable positions}.
    \label{def:acceptable}
         \end{defn}
The above definition can be extending
        to the entire class of
        probability measures on $(\Omega, \mc{F})$
as follows:
\begin{equation}
    \mc{A}=\cap_{P\in \mc{M}_1(\Omega, \mc{F})} \{X\in \mc{X}|\
    \mb{E}_P[\uv(X)]\geq \uv(y_{P})\}
    \label{eq:acceptance2}
\end{equation}
with constants $y_{P}$ such that $\sup_{P\in \mc{M}_1(\Omega, \mc{F})}y_{P}<\infty$.

\begin{prop}[{\cite[Proposition~4.7]{follmer:2002aa}}]
Suppose the class of \emph{acceptable positions}  $\mc{A}$ is a
non-empty subset of $\mc{X}$ satisfying
\begin{enumerate}
\item $\inf\{m\in \mb{R}| X+m\in \mc{A}\}>-\infty$, $\forall \ X\in
             \mc{X}$, and
         \item   given $X\in \mc{A}, Y\in \mc{X}$,  $Y\geq X$ $\Longrightarrow$ $Y\in
             \mc{A}$.
\end{enumerate}
Then, $\mc{A}$ induces a monetary measure of risk $\rho_{\mc{A}}$. If $\mc{A}$ is convex,
then $\rho_{\mc{A}}$ is a convex measure of risk.   
 Furthermore, if $\mc{A}$ is
     a cone, then $\rho_{\mc{A}}$ is a coherent risk metric.
\end{prop}
Note that a
monetary measure of risk induced by  a set of acceptable positions
$\mc{A}\subset \mc{X}$ is given by
\begin{equation}
    \rho_{\mc{A}}(X)=\inf\{z\in \mb{R}|\ z+X\in \mc{A}\}.
    \label{eq:rhoAA}
\end{equation}

The following proposition is key for extending the expectation operator to more
general measures of risk.
\begin{prop}[{\cite[Proposition~4.104]{follmer:2002aa}}]
         Consider 
     \begin{equation}
        \mc{A}=\{X\in \mc{X}|\ \mb{E}_P[\uv(X)]\geq \uv_0\}
        \label{eq:acceptablepos}
    \end{equation}
    for a continuous value function $\uv$, acceptance level $\uv_0=v(\bar{y})$ for
    some $\bar{y}$ in the domain of $v$, and probability measure $P$. Suppose
    that $\uv$ is strictly increasing
    on $(\bar{y}-\vep,\infty)$ for some $\vep>0$. Then, the corresponding $\rho_{\mc{A}}$ is a
    convex measure of risk which is continuous from below. Moreover,
    $\rho_{\mc{A}}(X)$ is the unique solution to 
    \begin{equation}
        \mb{E}_P[\uv(X-m)]= \uv_0.
        \label{eq:eqcond}
    \end{equation}
    \label{prop:optimality}
     \end{prop}
     Proposition~\ref{prop:optimality} also implies that for each value function, we can define an acceptance set which in turn
     induces a convex risk metric $\rho$. 
     Let us consider an example.
    \begin{example}[Entropic Risk Metric~\cite{follmer:2002aa}]
        Consider the entropic value function $\uv(y)=\exp(\lambda y)$. It has been used extensively in the field of risk
         measures~\cite{follmer:2002aa}, in neuroscience to capture risk
         sensitivity in motor control~\cite{nagengast:2010aa} and even more so in control of
         MDPs (see,
         \emph{e.g.},~\cite{coraluppi:2000aa}). 
        
        The entropic value function with an acceptance level $v_0$ can be used to define the acceptance set
        \begin{equation}
            \mc{A}=\{m\in \mb{R}|\ \mb{E}[\exp(-\lambda(m+Y))]\leq v_0\}.
            \label{eq:acept}
        \end{equation}
         with corresponding risk metric  
         \begin{align}
             \rho_{\mc{A}}(X)&=\inf\{m\in \mb{R}|\ \mb{E}[\exp(-\lambda(m+Y))]\leq v_0\}\\
             &=\textstyle\frac{1}{\lambda}\log
             \mb{E}[\exp(-\lambda Y)]-\frac{1}{\lambda}\log(v_0).
             \label{eq:entropic}
         \end{align}
The parameter $\lambda\in \mb{R}$ controls
         the risk preference; indeed, this can be seen by considering the Taylor
         expansion~\cite[Example~4.105]{follmer:2002aa}. 

         As a further comment,
         this particular risk metric is equivalent (up to an additive constant)
         to the so called \emph{entropic risk measure} which is given by
         \begin{equation}
             \rho(Y)=\sup_{P'\in \mc{M}_1(P)}\left(
             \mb{E}_{P'}[-Y]-\frac{1}{\lambda}H(P'|P) \right)
             \label{eq:entropicrisk2}
         \end{equation}
         where $\mc{M}_1(P)$ is the set of all measures on $(\Omega, \mc{F})$
         that are absolutely continuous with respect to $P$ and where
         $H(\cdot|\cdot)$ is the relative entropy function.
         \hfill\ensuremath{\blacksquare}
          \label{ex:entropic}
\end{example}

Let us
     recall the concept of
     a \emph{valuation function} introduced and used in~\cite{follmer:2002aa,
shen:2014aa,artzner:1999aa}.
     \begin{defn}[Valuation Function]
    A mapping $\mc{V}: \mb{R}^{|I|}\times \Delta(I)\rar \mb{R}$ is called a
    \emph{valuation function} if for each $\mu\in \Delta(I)$, (i) 
    $\mc{V}(Y,\mu)\leq \mc{V}(Z, \mu)$ whenever $Y\leq Z$ (monotonic) and (ii)
    $\mc{V}(Y+y\mathbf{1},\mu)=\mc{V}(Y,\mu)+y$ for any $y\in \mb{R}$ (translation
    invariant).
\end{defn}
Such a map is used to characterize an agent's preferences---that is, one prefers $(Y, \mu)$ to $(Z, \nu)$ whenever
 $\mc{V}(Z, \nu)\leq \mc{V}(Y, \mu)$.

 We will consider valuation functions that are convex risk metrics induced by a
 value function $\uv$ and a probability measure $\mu$.
 To simplify notation, from here on out we will suppress the dependence on the probability
measure $\mu$.

     For each state--action pair, we define     
     $\mc{V}(Y|x,a):\mb{R}^{|I|}\times X\times A\rar \mb{R}$ a \emph{valuation
     map} such that $\mc{V}_{x,a}\equiv \mc{V}(\cdot|x,a)$ is a valuation
     function induced by an acceptance set with respect to value function $\uv$
     and acceptance level $v_0$.

      If we let $\mc{V}_x^\pi(Y)=\sum_{a\in A(x)}\pi(a|x)\mc{V}_{x,a}(Y)$, 
     the optimization problem in \eqref{eq:exputil} generalizes to
       \begin{align}
           \textstyle\tilde{J}_{T}(\pi,x_0)=&\textstyle\mc{V}_{x_0}^{\pi_0}\Big[
               R[x_0,a_0]+\gamma\mc{V}_{x_1}^{\pi_1}\big[R[x_1,a_1]+
               \cdots+\gamma\mc{V}_{x_T}^{\pi_T}[R(x_T,a_T)]\cdots
           \big]\Big]
         \label{eq:genopt}
     \end{align}
    where we  define
     $\max_{\pi}\tilde{J}(\pi,x_0)=\lim_{T\rar\infty}\tilde{J}_T(\pi,x_0)$.
     
     
     
     
     \subsection{Risk-Sensitive Q-Learning Convergence}
     In the classical {reinforcement learning} framework, the Bellman equation is used to derive a
     Q-learning procedure. Generalizations of the Bellman equation for
     risk-sensitive {reinforcement learning}---derived, \emph{e.g.},  in
     \cite{shen:2013aa,mihatsch:2002aa}---have been
used to formulate an
     action--value function or Q-learning procedure for the risk-sensitive
     reinforcement learning
     problem.
     In particular, as shown in~\cite{shen:2013aa}, if $V^\ast$ satisfies 
          \begin{equation}
         V^\ast(x_0)=\textstyle\max_{a\in A(x)}\mc{V}_{x,a}(R(x,a)+\gamma V^\ast),
         \label{eq:riskBellman}
     \end{equation}
     then
         $V^\ast=\max_\pi\tilde{J}(\pi,x_0)$ holds for all $x_0\in X$; moreover,
          a deterministic policy is optimal if $\pi^\ast(x)=\arg\max_{a\in
         A(x)}\mc{V}_{x,a}(R+\gamma V^\ast)$~{\cite[Thm.~5.5]{shen:2013aa}}.
The action--value function $Q^\ast(x,a)=\mc{V}_{x,a}(R+\gamma
V^\ast)$ is defined such that \eqref{eq:riskBellman} becomes
\begin{equation}
    Q^\ast(x,a)=\textstyle\mc{V}_{x,a}\left(R+\gamma
    \max_{a\in A(x')}Q^\ast(x',a)\right),    \label{eq:Qstar}
\end{equation}
for all $(x,a)\in X\times A$.

Given a value function $\uv$ and acceptance level $\uv_0$, we use the coherent risk metric
induced state-action valuation function given by
\begin{equation}
    \mc{V}_{x,a}(Y)=\sup\{ z\in \mb{R}|\ \mb{E}[\uv(Y-z)]\geq \uv_0\}
    \label{eq:stateactionvaluation}
\end{equation}
 where the
expectation is taken with respect to $\mu=P(x'|x,a)P_r(w|x,a)$. Hence, by a direct
application of Proposition~\ref{prop:optimality}, if $\uv$ is
continuous and strictly increasing, then $\mc{V}_{x,a}(Y)=z^\ast(x,a)$ is the unique solution
to $\mb{E}[\uv(Y-z^\ast(x,a))]=\uv_0$.

As shown in~\cite[Proposition~3.1]{shen:2014aa}, by
letting $Y=R+\gamma V^\ast$, we have that $z^\ast(x,a)$ corresponds to
$Q^\ast(x,a)$ and, in particular,
\begin{align}
    \mb{E}\left[ \uv\left( r(x,a,w)+\gamma \max_{a'\in
    A(x')}Q^\ast(x',a')-Q^\ast(x,a) \right) \right]=\uv_0
    \label{eq:optimality}
\end{align}
where, again, the expectation is taken with respect to $\mu=P(x'|x,a)P_r(w|x,a)$.

 The above leads naturally to a Q-learning procedure,
\begin{align}
    Q(x_t,a_t)\leftarrow&Q(x_t,a_t)+\alpha_t(x_t,a_t)\big[\uv(y_t)-\uv_0\big],
    \label{eq:Qlearnproc}
\end{align}
where the non-linear transformation $\uv$  is applied to the temporal
difference \[y_t=r_t+\gamma \max_a
Q(x_{t+1},a)-Q(x_t,a_t)\] instead of simply the reward $r_t$.
Transformation of the temporal
differences avoids certain pitfalls of the reward transformation
approach such as poor convergence performance. This procedure has convergence guarantees even in this more general setting
under some assumptions on the value function $\uv$. 



\begin{thm}[Q-learning Convergence~{\cite[Theorem~3.2]{shen:2014aa}}]
    Suppose that $\uv:Y\rar \mb{R}$ is in $C(Y, \mb{R})$, is strictly increasing
    in $y$ and there exists constants $\vep, L>0$ such that $\vep\leq
    \frac{\uv(y)-\uv(y')}{y-y'}\leq L$ for all $y\neq y'$. Moreover, suppose
    that there exists a $\bar{y}$ such
    that $v(\bar{y})=\uv_0$. If the non-negative learning rates $\alpha_t(x,a)$ are such
that
$\sum_{t=0}^\infty \alpha_t(x,a)=\infty$ and $\sum_{t=0}^\infty
\alpha_t^2(x,a)<\infty$, $\forall(x,a)\in X\times A$, then the procedure in
\eqref{eq:Qlearnproc} converges to $Q^\ast(x,a)$ for all $(x,a)\in X\times A$
with probability one.
\label{thm:qlearn}
\end{thm}
The assumptions on $\alpha_t$ are fairly standard and the core of the
convergence proof is
based on the Robbins--Siegmund Theorem appearing in the seminal work~\cite{robbins:1985aa}.

We note that the assumptions on the value function $v$ of
Theorem~\ref{thm:qlearn} are fairly restrictive, excluding many of the value
functions presented in Section~\ref{sec:value}. For example, value functions of
the form $e^x$ and $x^\zeta$ do not satisfy the global Lipschitz condition.

We generalize the convergence result in Theorem~\ref{thm:qlearn} by modifying the assumptions on the value function
$v$ to ensure that we have convergence of the Q-learning procedure for the
{\lprospect} and entropic value functions.
\begin{ass}
    The value function $\uv\in
  C^1(Y, \mb{R})$ satisfies the following:
    \begin{enumerate}
        \item[(i)]  it is strictly
            increasing in $y$ and  there exists a $\bar{y}$ such
    that $\uv(\bar{y})=\uv_0$;
\item[(ii)]  it is locally Lipschitz on any ball of finite radius centered at
    the origin;
    \end{enumerate}
    \label{ass:v}
\end{ass}
Note that in comparison to the assumptions of Theorem~\ref{thm:qlearn}, we have
removed the assumption that the derivative of $v$ is bounded away from zero, and
relaxed the global Lipschitz assumption on $v$.
We remark that the {\lprospect} and entropic value functions satisfy these assumptions
for all parameters and MDPs.

Let $\mc{X}$
be a complete metric space endowed with the $L_\infty$ norm and let
$\mc{Q}\subset \mc{X}$ be the space of maps $Q:X\times A\rar \mb{R}$.
Further, define $\tilde{\uv}\equiv \uv-\uv_0$.
We then re-write the $Q$--update equation in the form 
    \begin{align}
     Q_{t+1}(x,a)&= \left( 1-\frac{\alpha_t}{\alpha}
        \right)Q_t(x,a)+\frac{\alpha_t}{\alpha}\big( \alpha
        (\uv(y_t)-\uv_0)+Q_t(x,a) \big)
        \label{eq:Qtup-2}
    \end{align}
    where 
$\alpha\in (0,\min\{L^{-1},1\}]$ and we have suppressed the dependence
    of $\alpha_t$ on $(x,a)$. This is a standard update equation form in, \emph{e.g.},
    the stochastic approximation algorithm
literature~\cite{robbins:1951aa,tsitsiklis:1994aa,kushner:2003aa}.
In addition, we define the map given by
\begin{align}
        (TQ)(x,a)=&\alpha \mb{E}_{x',w}\big[
        \tilde{
        \uv}\big(r(x,a,w)+\gamma \max_{a'\in
    A}Q(x',a')-Q(x,a)\big)\big]+Q(x,a)
        \label{eq:Tupdate-1}
    \end{align}
which we will prove is a contraction.

\begin{thm}
    Suppose that $v$ satisfies Assumption~\ref{ass:v} and that for each
    $(x,a)\in X\times A$ the reward
    $r(x,a,w)$ is bounded almost surely---that is, there exists $0<M<\infty$
    such that $|r|<M$ almost surely.
    Moreover, let $\alpha\in (0,\min\{1,L^{-1}\}]$, for $L$, the Lipschitz
    constant of $v$ on $B_K(0)$. 
    \begin{enumerate}
        \item[(a)]  Let
    $B_K(0)\subset \mc{Q}$ be a closed ball of radius $K>0$ centered at zero.
    Then, $T:\mc{Q}\rar\mc{X}$ is a
    contraction. 
\item[(b)]Suppose $K$ is chosen such that
    \begin{equation}
        \frac{\max\{|\tilde{v}(M)|,
        |\tilde{v}(-M)|\}}{(1-\gamma)}<K\min_{y\in
        I_K}D\tilde{v}(y)\label{eq:boundass}\end{equation}
    where $I_K=[-M-K, M+K]$.
        Then, $T$ has a unique fixed point in $B_K(0)$.
    \end{enumerate}
           \label{thm:contract}
\end{thm}
The proof of the above theorem 
is provided in Appendix~\ref{app:thmContract}.

The following proposition shows that the {\lprospect} and entropic value 
functions satisfy the assumption in \eqref{eq:boundass}. Moreover, it shows that
the value functions which satisfy Assumption~\ref{ass:v} also satisfy
\eqref{eq:boundass}. 
\begin{prop}
    Consider a MDP with reward $r:X\times A\times W\rar \mb{R}$ bounded almost surely by $M$ 
    and
    $\gamma\in (0,1)$ and consider the condition
\begin{equation}
        \frac{\max\{|\tilde{v}(M)|,
        |\tilde{v}(-M)|\}}{(1-\gamma)}<K\min_{y\in
        I_K}D\tilde{v}(y).\label{eq:boundass-lprospect-1}\end{equation}
    \begin{enumerate}
        \item Suppose $v$ satisfies Assumption~\ref{ass:v} and that for some
            ${\vep}>0$,
            ${\vep}<\frac{v(y)-v(y')}{y-y'}$ for all $y\neq y'$. Then
            \eqref{eq:boundass-lprospect-1} holds.
        \item Suppose
    $v$ is an {\lprospect} value function with arbitrary parameters
    $(k_-,k_+, \zeta_-,\zeta_+)$ satisfying Assumption~\ref{ass:v}. Then there exists a $K$
    such that the {\lprospect}
    value function satisfies \eqref{eq:boundass-lprospect-1}.
     \item Suppose  that $v$ is an entropic value
    function. Then there exists a $C>0$ such that for any $|\lambda|\in (0,C)$
    where $v$ satisfies Assumption~\ref{ass:v}, \eqref{eq:boundass-lprospect-1}
    holds with $K=(\lambda)^{-1}$.
    \end{enumerate}
    \label{cor:valuefunctions}
\end{prop}
With Theorem~\ref{thm:contract} and Proposition~\ref{cor:valuefunctions}, we
can prove convergence of Q-learning for risk-sensitive reinforcement learning.
\begin{thm}[Q-learning Convergence on $B_K(0)$]
       Suppose that $v$ satisfies Assumption~\ref{ass:v} and that for each
   $(x,a)\in X\times A$ the reward
    $r(x,a,w)$ is bounded almost surely---that is, there exists $0<M<\infty$
    such that $|r|<M$ almost surely.
    Moreover, suppose the ball $B_K(0)$ is 
    chosen such that \eqref{eq:boundass} holds. If the non-negative learning rates $\alpha_t(x,a)$ are such
that
$\sum_{t=0}^\infty \alpha_t(x,a)=\infty$ and $\sum_{t=0}^\infty
\alpha_t^2(x,a)<\infty$, $\forall(x,a)\in X\times A$, then the procedure in
\eqref{eq:Qlearnproc} converges to $Q^\ast \in B_K(0)$
with probability one.
\label{thm:qlearnconvergence}
\end{thm}
Given Theorem~\ref{thm:contract}, the proof of
Theorem~\ref{thm:qlearnconvergence} follows directly the same proof as provided
in~\cite{shen:2013aa}. The key aspect of the proof is combining the fixed point
result of Theorem~\ref{thm:contract} with Robbins--Siegmund
Theorem~\cite{robbins:1985aa}.

Theorem~\ref{thm:contract},
Proposition~\ref{cor:valuefunctions}, and Theorem~\ref{thm:qlearnconvergence} extend the results for risk-sensitive
reinforcement learning presented in~\cite{shen:2013aa} by relaxing the
assumptions on the value functions for which the Q-learning procedure converges.


\section{Inverse Risk-Sensitive Reinforcement Learning}
\label{sec:rsirl}
We formulate the inverse risk-sensitive  reinforcement learning problem  as follows. First,
we select a parametric class of policies, $\{\pi_\theta\}_\theta$, $\pi_\theta \in
\Pi$ and parametric value function $\{\uv_\theta\}_\theta$, $\uv_\theta\in \mc{F}$
where $\mc{F}$ is a family of value functions and
$\theta\in \Theta\subset \mb{R}^d$. 

We use value functions such as those described in Section~\ref{sec:dm};
e.g., if $\uv$ is the prospect theory value function defined in
\eqref{eq:prospectu}, then the parameter vector is
$\theta=(\kl,\ku, \Ll,\Lu, \gamma, \beta)$. For mappings $\uv$ and $Q$,
we now indicate their dependence on $\theta$---that is, we will write
$Q(x,a,\theta)$ and $\uv_\theta(y)=\uv(y,\theta)$ where $\uv: Y\times
\Theta\rar\mb{R}$.  Note that since $y$ is the temporal difference it also
depends on $\theta$ and we will indicate this dependence where it is not
directly obvious by writing $y(\theta)$.

It is common in the {\IRL} literature to adopt a smooth map $G$ that
operates on the action-value function space for defining the parametric policy
space---{e.g.}, Boltzmann policies of the form 
\begin{equation}
   G_\theta(Q)(a|x)=\frac{\exp(\beta Q(x,a,\theta))}{\sum_{a'\in A}\exp(\beta
Q(x,a', \theta))}
    \label{eq:boltz}
\end{equation}
to the action-value functions $Q$ where $\beta>0$ controls how close $G_\theta(Q)$ is
to a \emph{greedy policy} which we define to be any policy $\pi$ such that $\sum_{a\in
A}\pi(a|x)Q(x,a,\theta)=\max_{a\in A}Q(x,a,\theta)$ at all states $x\in X$. 
We will utilize policies of this form. Note that, as is pointed out
in~\cite{neu:2007aa}, the benefit of selecting strictly stochastic policies is
that if the true agent's  policy is deterministic, uniqueness of
the solution is forced.

We aim to \emph{tune} the parameters so as
to minimize some loss $\ell(\pi_\theta)$ which is a function of the
parameterized policy
$\pi_\theta$. By an abuse of notation, 
we introduce the shorthand $\ell(\theta)=\ell(\pi_\theta)$.

\subsection{Inverse Reinforcement Learning Optimization Problem}
The optimization problem is specified by
\begin{equation}
\min_{\theta\in \Theta}\left\{\ell(\theta) | \
        \pi_\theta=G_\theta(Q^\ast), \uv_\theta\in \mc{F}\right\}
    \label{eq:optprob}
\end{equation}
Given a set of \emph{demonstrations} $\mc{D}=\{(x_k,a_k)\}_{k=1}^N$, it is our
goal to recover the policy and estimate the value function.

There are several possible loss functions that may be employed.
For example, suppose we elect to minimize the negative weighted
log-likelihood of the demonstrated behavior which is given by
\begin{equation}
      \ell(\theta)= \textstyle\sum_{(x,a)\in \mc{D}}w(x,a)\log(\pi_{\theta}(x,a))
    \label{eq:loglike}
\end{equation}
where $w(x,a)$ may, {e.g.}, be the normalized empirical frequency of
observing $(x,a)$ pairs in $\mc{D}$, i.e.~$n(x,a)/N$ where $n(x,a)$ is the
frequency of $(x,a)$.  

Related to maximizing the log-likelihood,
an alternative loss function is the relative entropy or
Kullback-Leibler (KL) divergence  between the empirical
distribution of the state-action trajectories and their distribution
under the learned policy---that is, 
\begin{equation}
  \ell(\theta)=\textstyle\sum_{x\in
        \mc{D}_x}D_{\text{KL}}(\hat{\pi}(\cdot|x)||\pi_\theta(\cdot|x))
    \label{eq:KLdivergence}
\end{equation}
where 
\begin{equation}
\textstyle    D_{\text{KL}}(P||Q)=\sum_iP(i)\log\left(P(i)/Q(i)\right)
    \label{eq:KLdivergence2}
\end{equation}
is the KL divergence,
 $\mc{D}_x\subset \mc{D}$ is the sequence of observed states, and
$\hat{\pi}$ is the empirical distribution on the trajectories of $\mc{D}$.

\subsection{Gradient--Based Approach }
\label{sec:irlgrad}
We propose to solve the problem of estimating the parameters of the agent's
value function and approximating the agent's policy via gradient methods which requires computing the
derivative of $Q^\ast(x,a,\theta)$ with respect to $\theta$.
Hence, given the form of the Q-learning procedure where the temporal
differences are transformed as in \eqref{eq:Qlearnproc}, we need to derive a mechanism for obtaining the
optimal $Q$, show that it is in fact differentiable, and derive a procedure for
obtaining the derivative.

Using some basic calculus, given the form of smoothing map $G_\theta$ in
\eqref{eq:boltz}, we can compute the
derivative of the policy $\pi_\theta$ with respect to $\theta_k$ for an element of
$\theta\in \Theta$:
\begin{align}
 D_{\theta_k} \pi_\theta(a|x)&=\textstyle\pi_\theta(a|x)D_{\theta_k}{\ln(\pi_\theta(a|x))}\\
    &= \textstyle \pi_\theta(a|x)\beta\big(
    D_{\theta_k}Q^\ast(x,a,\theta)-\sum_{a'\in
        {A}}\pi_\theta(a'|x)D_{\theta_k}Q^\ast(x,a',\theta) \big).
    \label{eq:pdpi}
\end{align}
We show that $D_{\theta_k}Q_\theta^\ast$ can be calculated almost
everywhere on $\Theta$ by solving fixed-point equations similar to the Bellman-optimality equations.

To do this, we require some assumptions on the value function $\uv$. 
\begin{ass}
    The value function $\uv\in
    C^1(Y\times \Theta, \mb{R})$
    satisfies the following
    conditions:
    \begin{enumerate}
        \item[(i)]  $v$ is strictly
            increasing in $y$ and for each $\theta\in \Theta$, there exists a $\bar{y}$ such
    that $\uv(\bar{y},\theta)=\uv_0$;
\item[(ii)] for each $\theta\in\Theta$, on any
    ball centered around the origin of finite radius, $v$ is locally Lipschitz in $y$  with constant
    $L_y(\theta)$
    and locally Lipschitz on $\Theta$ with constant $L_\theta$;
\item[(iii)] there exists $\vep>0$ such that $\vep\leq
    \frac{\uv(y,\theta)-\uv(y',\theta)}{y-y'}$ for all $y\neq y'$.
    \end{enumerate}
    \label{ass:u}
\end{ass}

Define $L_y=\max_\theta L_y(\theta)$ and
$L=\max_\theta\{L_y(\theta),L_\theta\}$. As before, let
 $\tilde{\uv}\equiv \uv-\uv_0$.
We re-write the $Q$--update equation as
 \begin{align}
     Q_{t+1}(x,a,\theta)&= \textstyle\left( 1-\frac{\alpha_t}{\alpha}
        \right)Q_t(x,a,\theta)+\frac{\alpha_t}{\alpha}\big( \alpha
        (\uv(y_t(\theta),\theta)-\uv_0)+Q_t(x,a,\theta) \big)
        \label{eq:Qtup-2}
    \end{align}
    where \[y_t(\theta)=r_t+\gamma\max_{a}Q_t(x_{t+1},a,\theta)-Q_t(x_t,a_t,
    \theta)\] is the temporal difference, $\alpha\in (0,\min\{L^{-1},1\}]$ and we have suppressed the dependence
    of $\alpha_t$ on $(x,a)$.
    In addition, define the map $T$ such
    that
    \begin{equation}
        (TQ)(x,a,\theta)=\alpha \mb{E}_{x',w}
        \tilde{
        \uv}(y(\theta),\theta)+Q(x,a, \theta)
        \label{eq:Tupdate}
    \end{equation}
    where $y(\theta)=r(x,a,w)+\gamma \max_{a'\in
    A}Q(x',a',\theta)-Q(x,a,\theta)$. 
This map is a contraction for each $\theta$. Indeed, fixing $\theta$,
when $v$ satisfies
Assumption~\ref{ass:u}, then for cases where $\uv_0=0$, $T$ was shown to be 
a contraction 
in~\cite{mihatsch:2002aa} and in the more general setting (i.e.~$\uv_0\neq 0$), 
in~\cite{shen:2014aa}.

Our first main result on inverse risk-sensitive reinforcement learning, which is the theoretical underpinning of our gradient-based algorithm, gives
us a mechanism to compute the derivative of $Q^\ast_\theta$ with respect to
$\theta$ as a solution to a fixed-point equation via a contraction mapping
argument.

Let $D_i\tilde{\uv}(\cdot,\cdot)$ be the
derivative of $\tilde{\uv}$ with respect to the $i$--th argument where $i=1,2$.
\begin{thm} 
    Assume that $\uv\in C^1(Y\times \Theta, \mb{R})$ satisfies Assumption~\ref{ass:u}.
    Then the following statements hold:
    \begin{enumerate}[(a)]
        \item $Q_\theta^\ast$ is locally Lipschitz continuous as a function of
            $\theta$---that is, for any $(x,a)\in {X}\times {A}$, $\theta,
            \theta'\in \Theta$, $|Q^\ast(x,a,\theta)-Q^\ast(x,a, \theta')|\leq
            C
            \|\theta-\theta'\|$ for some $C>0$;
        \item except on a set of measure zero, the gradient $D_\theta
            Q_\theta^\ast$ is given by the solution of the  fixed--point equation
              \begin{align}
           \phi_\theta(x,a)=&\alpha \mb{E}_{x',w}\big[ D_2
           \tilde{\uv}(y(\theta),\theta)+D_1\tilde{\uv}(y(\theta), \theta)(\gamma
       \phi_\theta(x',a_{x'}^\ast)-\phi_\theta(x,a)) \big]+\phi_\theta(x,a)
           \label{eq:Scontraction}
       \end{align}
       where $\phi_\theta:X\times A\rar \mb{R}^d$ and $a_{x'}^\ast$  is the action that maximizes $\sum_{a'\in
       A}\pi(a|x')Q(x',a,\theta)$ where $\pi$ is any policy that is greedy with
       respect to $Q_\theta$.
    \end{enumerate}
    \label{thm:DQ}
\end{thm}
We provide the proof in Appendix~\ref{app:thmDQ}. To give a high-level outline,
we use an induction argument combined with a contraction mapping argument on the
map
\begin{align}
     (S\phi_\theta)(x,a)
           =&\alpha \mb{E}_{x',w}\big[ D_2
           \tilde{\uv}(y(\theta),\theta)+D_1\tilde{\uv}(y(\theta),
           \theta)(\gamma
       \phi_\theta(x',a_{x'}^\ast)-\phi_\theta(x,a)) \big]+\phi_\theta(x,a).
    \label{eq:Scontraction-1}
\end{align}
 The almost everywhere differentiability follows from
Rademacher's Theorem (see, \emph{e.g.},~\cite[Thm.~3.1]{heinonen:2004aa}).

Theorem~\ref{thm:DQ} gives us a procedure---namely, a fixed--point equation
which is a contraction---to compute the derivative $D_{\theta_k}Q^\ast$ so that we can compute the derivative of our loss function
$\ell(\theta)$. Hence the gradient method provided in Algorithm~\ref{alg:gbrsirl}
for solving the inverse risk-sensitive reinforcement learning problem is well formulated.

\begin{algorithm}
\caption{Gradient-Based Risk-Sensitive IRL}\label{alg:gbrsirl}
\begin{algorithmic}[1]
    \Procedure{RiskIRL}{$\mc{D}$} 
\State Initialize: $\theta\leftarrow\theta_0$
\While{$k< \text{MAXITER}$ \& $\|\ell(\theta)-\ell(\theta_{-})\|\geq
\delta$} 
\State $\theta_{-}\gets \theta$
\State $\eta_k\gets$ \Call{LineSearch}{$\ell(\theta_{-}),D_\theta
\ell(\theta_{-})$}
\State $\theta\leftarrow \theta_{-}-\eta_k D_\theta \ell(\theta_{-})$
\State $k\gets k+1$
\EndWhile\label{euclidendwhile}
\State \textbf{return} $\theta$ 
\EndProcedure
\end{algorithmic}
\end{algorithm}

\begin{remark}
    The prospect theory value function $\uv$ given in \eqref{eq:prospectu} is not globally
    Lipschitz in $y$---in particular, it is not Lipschitz near the reference
point $\rp$---for values of $\Lu$ and $\Ll$  less than one. Moreover,
for certain parameter combinations, it may not even be differentiable. The
{\lprospect} function, on the other hand, is locally Lipschitz and its
derivative near the reference point is bounded away from zero. This makes it a
more viable candidate for numerical implementation. Its derivative,
however, is not bounded away from zero as $y\rar \infty$. 

This being said, we note that if the procedure for computing $Q^\ast$
 follows an algorithm which implements repeated
applications of the map $T$ is initialized with $Q_0(x,a)$ being finite for all
$(x,a)$ and $r$ is bounded for all possible $(x,a,w)$ pairs, then the derivative
of $\tilde{\uv}$ will always be bounded away from zero for all realized
values of $y$ in the procedure. 
An analogous statement can be made regarding the computation of $D_\theta Q^\ast$.
Hence, the procedures for computing $Q^\ast$ and $D_\theta Q^\ast$ for all the value functions we consider (excluding the classical prospect
value function) are guaranteed to converge (except on a set of measure zero).
\end{remark}

Let us translate this remark into a formal result. Consider a modified version
of Assumption~\ref{ass:u}:
\begin{ass}
    The value function $\uv\in C^1(Y\times \Theta, \mb{R})$ satisfies the following:
    \begin{enumerate}
        \item[(i)]  it is strictly
            increasing in $y$ and for each $\theta\in \Theta$, there exists a $\bar{y}$ such
    that $\uv(\bar{y},\theta)=\uv_0$;
\item[(ii)] for each $\theta\in\Theta$, it is Lipschitz in $y$ with constant
    $L_y(\theta)$
    and locally Lipschitz on $\Theta$ with constant $L_\theta$.
    \end{enumerate}
    \label{ass:uv}
\end{ass}
Simply speaking, analogous to Assumption~\ref{ass:v}, we have removed the uniform lower bound on the derivative of
$\uv$. Moreover, Theorem~\ref{thm:contract} gives us that $T$, as defined
in~\eqref{eq:Tupdate}, 
is a contraction on a ball of finite radius for each $\theta$ under Assumption~\ref{ass:v}.

\begin{thm}
        Assume that $\uv\in C^1(Y\times \Theta, \mb{R})$ satisfies
        Assumption~\ref{ass:uv} and that the reward $r: X\times A\times
        W\rar\mb{R}$ is bounded almost
        surely by $M>0$.
        Then the following statements hold.
    \begin{enumerate}[(a)]
        \item  For any ball $B_K(0)$, $Q_\theta^\ast$ is locally Lipschitz-continuous on $B_K(0)$ as a function of
            $\theta$---that is, for any $(x,a)\in {X}\times {A}$, $\theta,
            \theta'\in \Theta$, $|Q^\ast(x,a,\theta)-Q^\ast(x,a, \theta')|\leq
            C
            \|\theta-\theta'\|$ for some $C>0$.
        \item  
        For each $\theta$, 
            let $B_K(0)$ be the
            ball with radius $K$ satisfying
            \begin{equation}
                \frac{\max\{|\tilde{v}(M,\theta)|,| \tilde{v}(-M,
            \theta)|\}}{1-\gamma}<K\min_{y\in I_K}.
        D\tilde{v}(y,\theta)
        \label{eq:vtildecond1}
    \end{equation}
            Except on a set of measure zero, the gradient $D_\theta
            Q_\theta^\ast(x,a)\in B_K(0)$ is given by the solution of the  fixed--point equation
              \begin{align}
           \phi_\theta(x,a)=&\alpha \mb{E}_{x',w}\big[ D_2
           \tilde{\uv}(y(\theta),\theta)+D_1\tilde{\uv}(y(\theta), \theta)(\gamma
       \phi_\theta(x',a_{x'}^\ast)-\phi_\theta(x,a)) \big]+\phi_\theta(x,a)
           \label{eq:Scontraction}
       \end{align}
       where $\phi_\theta:X\times A\rar \mb{R}^d$ and $a_{x'}^\ast$  is the action that maximizes $\sum_{a'\in
       A}\pi(a|x')Q(x',a,\theta)$ with $\pi$ being  any policy that is greedy with
       respect to $Q_\theta$.
    \end{enumerate}
    \label{thm:DQball}
\end{thm}
The proof (provided in Appendix~\ref{app:thmDbound})  of the above theorem
follows the same techniques as in Theorem~\ref{thm:contract} and
Theorem~\ref{thm:DQ}.

Note that for each fixed $\theta$, condition~\eqref{eq:vtildecond1} is the same as
condition~\eqref{eq:boundass}. Moreover, Proposition~\ref{cor:valuefunctions}
shows that for the
{\lprospect}  and entropic value functions, such a $K$ must exist for any choice
of parameters.

\subsection{Complexity}
\label{sec:complexity}

Small dataset size is often a challenge in modeling sequential human
decision-making owing in large part to the frequency and time scale on which decisions are
made in many applications. To properly understand how our gradient-based approach performs for different amounts of data, we analyze the case when the loss function,
$\ell(\theta)$, is either the negative of the log-likelihood of
the data---see  \eqref{eq:loglike} above---or the sum over states of the KL divergence between the
policy under our learned value function and the the empirical policy of the
agent---see \eqref{eq:KLdivergence} above. These are two of the
more common loss functions used in the literature.

We first note that maximizing the log-likelihood is equivalent to minimizing a weighted sum over
states of the KL divergence between the empirical policy of the \emph{true} agent,
$\hat\pi_n$,
and the policy under the learned value function, $\pi_\theta$. In particular,
through some algebraic manipulation the weighted log-likelihood 
can be re-written as
\begin{equation}
      \ell(\theta)=\textstyle \sum_{x\in
        \mc{D}_x} w(x) D_{KL}(\hat\pi_{n}(\cdot|x)||\pi_{\theta}(\cdot|x))
\end{equation}
where $w(x)$
is the frequency of state $x$ normalized by $|\mc{D}|=N$. This approach has the added benefit
that it is independent of $\theta$ and therefore will not be affected by scaling
of the value functions~\cite{neu:2007aa}. 

Both cost functions are natural metrics for performance in that they minimize a
measure of the divergence between the optimal policy under the learned agent and
empirical policy of the true agent. While the KL-divergence is not suitable
for our analysis, since it is not a metric on the space of probability
distributions,
it does provide an upper bound on the total
variation (TV) distance via Pinsker's inequality:
\begin{equation}
      \delta(\hat\pi_n(\cdot|x),\pi_\theta(\cdot|x))\le \sqrt{2D_{KL}(\hat\pi_n(\cdot|x)||\pi_\theta(\cdot|x))}
      \label{eq:Pinsker}
\end{equation}
where $\delta(\pi(\cdot|x),\pi_\theta(\cdot|x))$ is the TV distance between $\hat\pi_n(\cdot|x)$ and $\pi_\theta(\cdot|x)$, defined as
\begin{equation}
    \delta(\hat\pi_n(\cdot|x),\pi_\theta(\cdot|x))=\textstyle\frac{1}{2}\|\pi_\theta(\cdot|x)-{\hat\pi_n}(\cdot|x)\|_1.
    \label{eq:TVnorm}
\end{equation}
The TV distance between distributions is a proper metric. Furthermore, use of the two cost functions described
above will also translate to minimizing the TV distance as it is upper
bounded by the KL divergence.

We first note that, for each state $x$, we would ideally like to get a bound on
$\delta(\pi(\cdot|x),\pi_\theta(\cdot|x))$, the TV distance between the agent's
true policy $\pi(\cdot|x)$ and the estimated policy $\pi_\theta(\cdot|x)$. 
However, we only have access to the empirical policy $\hat{\pi}_n$.
We therefore use the triangle inequality to get an upper bound on 
$\delta(\pi(\cdot|x),\pi_\theta(\cdot|x))$, in terms of values for which we can calculate explicitly or construct bounds.
In particular, we derive the following bound:
\begin{equation}
    \delta(\pi_\theta(\cdot|x),\pi(\cdot|x))\le 
    \delta(\hat{\pi}_n(\cdot|x),\pi_\theta(\cdot|x)) +
    \delta(\hat\pi_n(\cdot|x),\pi(\cdot|x)).
\label{eq:bound1}
\end{equation}
Note that $\delta(\hat{\pi}_n(\cdot|x),\pi_\theta(\cdot|x))$ is tantamount to
a training error as metricized by the TV distance, and is upper bounded by a
function of the KL divergence (which appears in the loss function) via \eqref{eq:Pinsker}.

The first term in \eqref{eq:bound1}, $\delta(\hat{\pi}_n(\cdot|x),\pi(\cdot|x))$, is the distance
between the empirical policy and the true policy in state $x$.  Using the Dvoretzky Kiefer-Wolfowitz inequality (see,
\emph{e.g.},~\cite{millar:1979aa,massart:1990aa}), this term can be bounded
above with high probability. Indeed,
\begin{equation}\text{Pr}(\|\pi(\cdot|x)-\hat{\pi}_n(\cdot|x)\|_1 > \epsilon) \le
    2|A|e^{-2n\epsilon^2/|A|^2}, \ \epsilon>0\end{equation}
where $n$ is the number of samples from the distribution $\pi(\cdot|x)$ and $|A|$ is the
cardinality of the action set.
Combining this bound with \eqref{eq:bound1}, we get  that, with probability $1-\nu$,
\begin{align}
  \delta(\pi_\theta(\cdot|x),\pi(\cdot|x))\le&
|A|\left(\frac{2}{n}\log\frac{2|A|}{\nu}\right)^{1/2}+
\delta(\hat{\pi}_n(\cdot|x),\pi_\theta(\cdot|x)).
\label{eq:bnd}
\end{align}

Supposing Algorithm 1 achieves a sufficiently small training error $\varepsilon>0$, 
the second term above can be bounded above by a calculable small amount which we
define notationally to be $\bar\varepsilon>0$. Supposing $\bar{\varepsilon}$ is also sufficiently small, 
the dominating term in the distance between $\pi$ and $\pi_\theta$ is the first
term on the right-hand side in \eqref{eq:bnd}. This gives us a $O(n^{-1/2})$
convergence rate on the per state level. This rate is seen qualitatively in our
experiments on sample complexity outlined in Section~\ref{sec:results_complex}.

We note that this bound is for each individual state $x$. Thus, for states that
are visited more frequently by the agent, we have better guarantees on how well
the policy under the learned value function approximates the true policy.
Moreover, it suggests ways of designing data collection schemes to better understand the agent's actions in less explored regions of the state space.


\section{Examples}
\label{sec:examples}

%
Let us now demonstrate the  performance of the proposed method on two examples. While
we are able to formulate the inverse risk-sensitive reinforcement learning problem for parameter vectors
$\theta$ that include $\gamma$ and $\beta$,  in the following examples 
 we use $\gamma=0.95$ and $\beta=4$. The purpose of doing this is to explore
 the effects of changing the value function parameters on the resulting policy.
 
 In all experiments, our optimization objective is the negative
 log-likelihood of the data, defined in \eqref{eq:loglike} and the valuation
 function we use is induced by an acceptance level set defined for a value
 function that we specify and acceptance level of zero. Furthermore, for the
 prospect and {\lprospect} value functions, we use a reference point of
 zero\footnote{Individually, the acceptance level and the reference point can be
 recentered around zero without loss of generality.}. These choices are aimed at
 further deconflating our observations of the behavior---in terms of
 risk-sensitivity and loss-aversion---that results from different choices of the value function
 parameters from other characteristics of the MDP or learning algorithm.



\subsection{Grid World}

In our first test of the proposed gradient-based inverse risk-sensitive
reinforcement learning approach, we utilize 
 data from agents operating on the canonical Grid World MDP. 
In the remainder, we describe the setup of the MDP, the three types of experiments we conduct,
and qualitative results on sample complexity. The three experiments are
described as follows:
\begin{enumerate}
    \item Learning the value function of an agent with the correct
model for the value function (e.g.,~learning a prospect value function when the
agent also has a prospect value function);
\item  Learning the value function of an agent with the wrong model
for the value function (e.g.,~learning an entropic map value function when the
agent has a prospect value function); 
\item Exploring the dependence of our training error on the number of sample trajectories collected from the agent.
\end{enumerate}
We measure the performance of the gradient-based approach via
the TV norm, defined in \eqref{eq:TVnorm}, of the difference between the policy in state $x$ of the {true} agent and the policy in state $x$ under the learned value function. 
\ifsc
\begin{figure*}[t]
    \centering
    \hfill\subfloat[][]{  \label{fig:gridsetup}
\includegraphics[width=0.375\columnwidth]{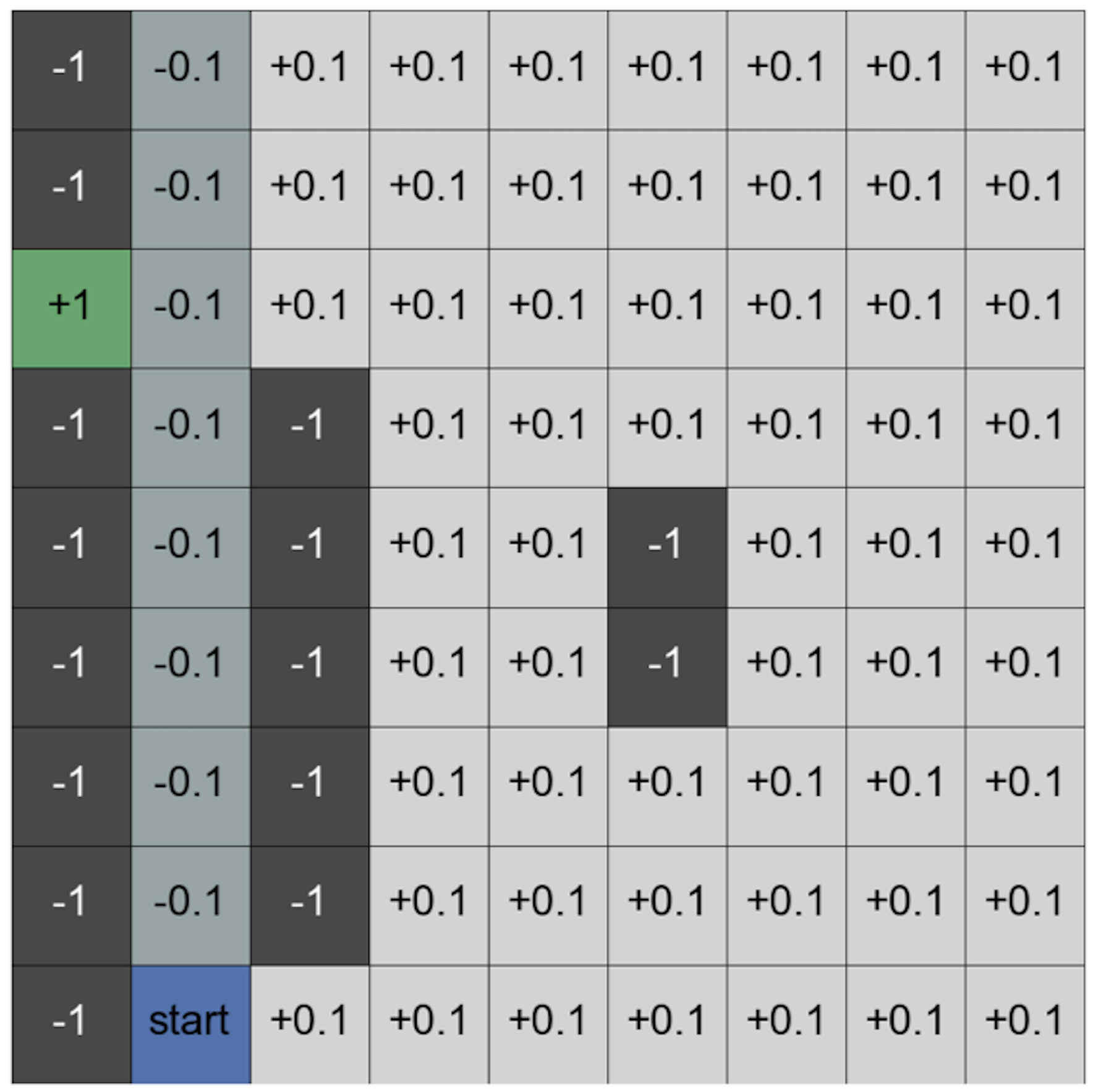}}
\hfill\subfloat[][]{ \label{fig:paths}
 \includegraphics[width=0.5075\columnwidth]{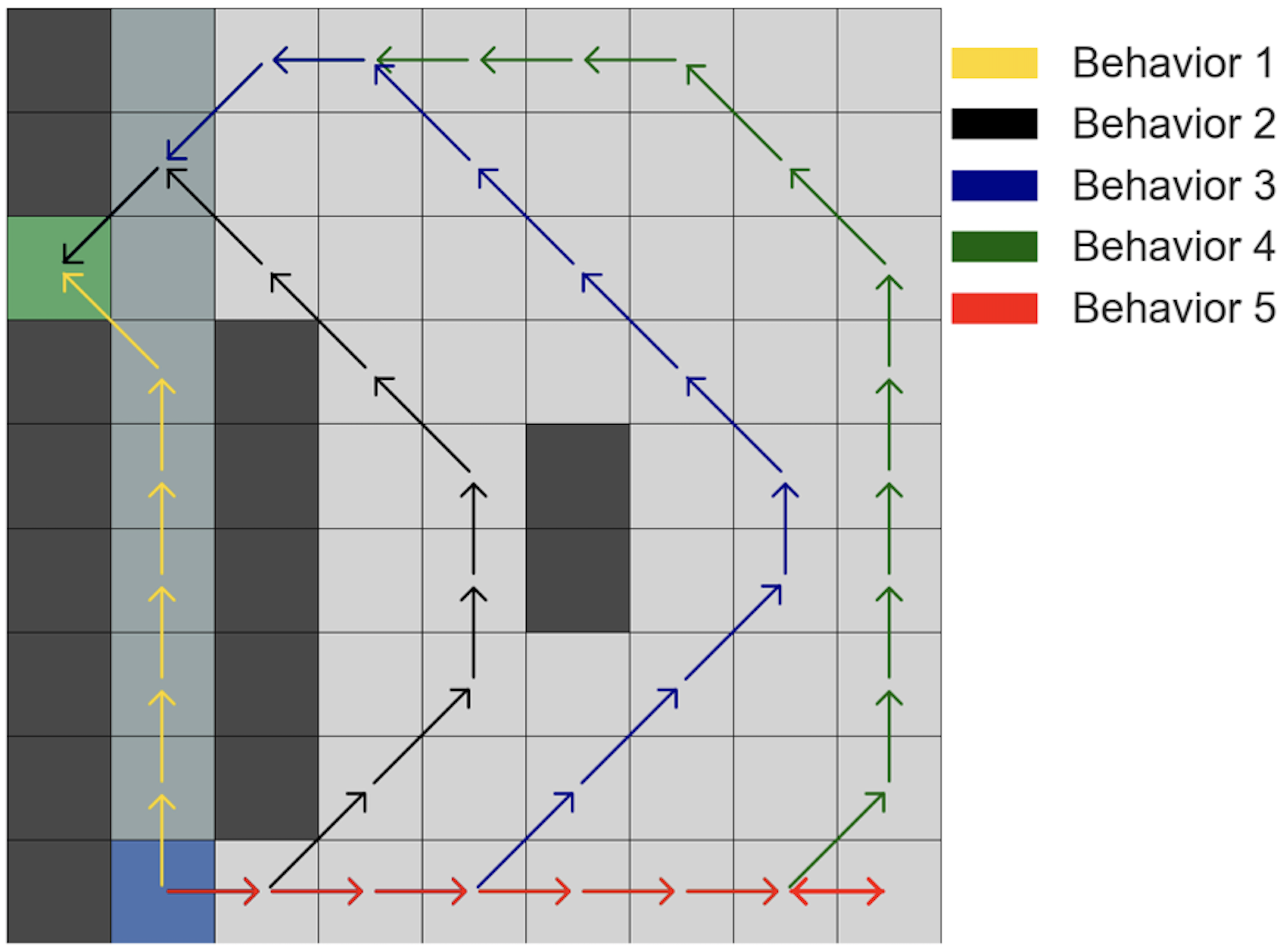}\hfill
}
\caption{(a) Grid World layout showing the reward structure. (b) {The five behavior profiles of risk-sensitive policies through the
        Grid World. These five paths correspond to the maximum likelihood paths of
        agents with various parameter combinations for their prospect,
        {\lprospect} and entropic map value functions. To generate each
        behavior with the prospect and {\lprospect} value functions, the following
        parameter combinations ($\{\kl,\ku,\Ll,\Lu\}$) were used: 
        \emph{Behavior 1}: $\{0.1,1.0,0.5,1.5\}$; \emph{Behavior 2}: $\{1.0,1.0,1.0,1.0\}$; \emph{Behavior
        3}: $\{1.0,1.0,1.1,0.9\}$; \emph{Behavior 4}: $\{5.0,1.0,1.1,0.8\}$; \emph{Behavior 5}: $\{5.0,1.0,1.5,0.7\}$.
    To generate the behaviors with the entropic map value function, we varied
    $\lambda$ from $1$ to $-1$.
}}
   \end{figure*}

\else
\begin{figure*}[t]
    \centering
    \subfloat[][]{  \label{fig:gridsetup}
\includegraphics[width=0.57\columnwidth]{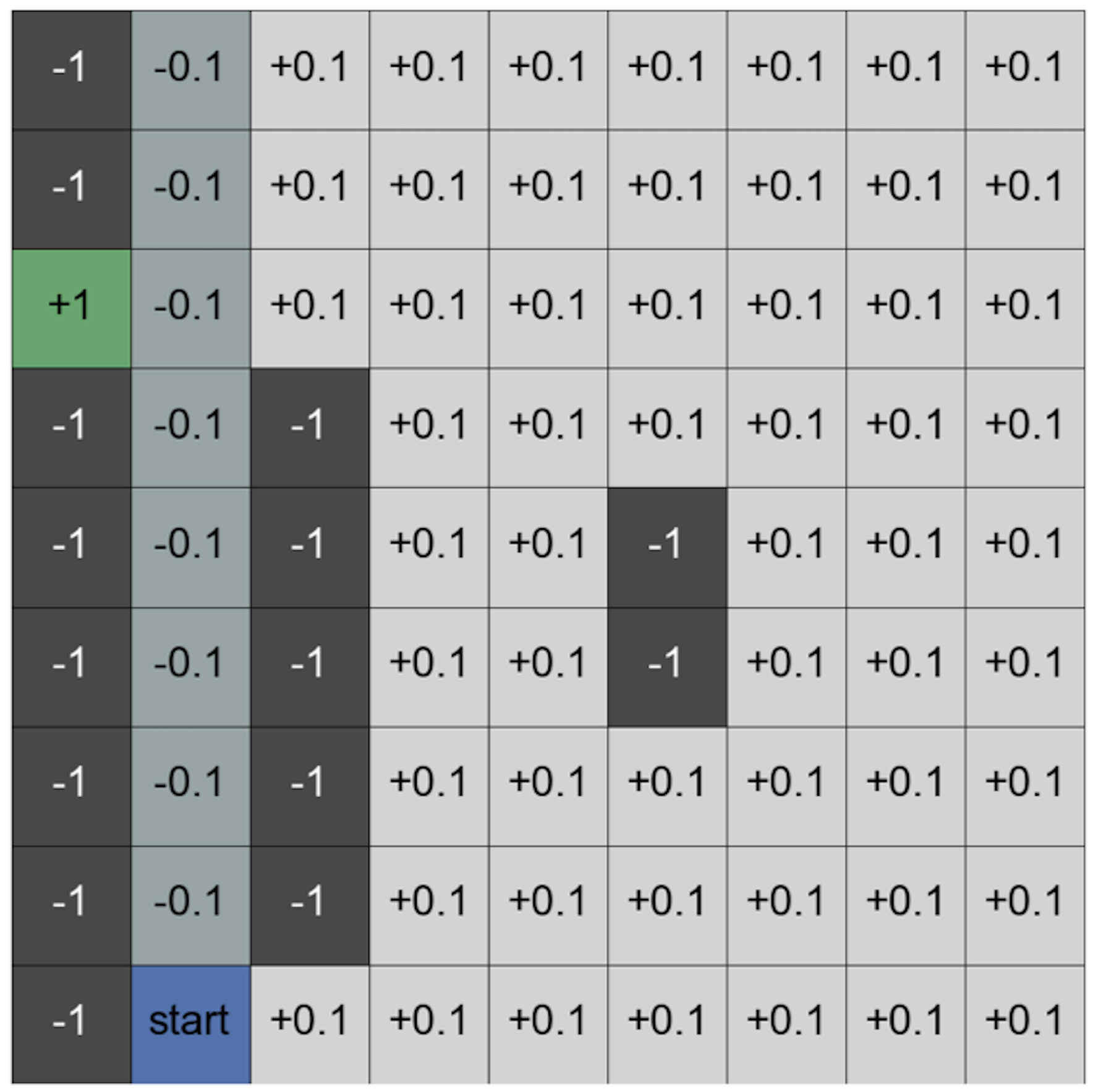}}
$\qquad\qquad$\subfloat[][]{ \label{fig:paths}
 \includegraphics[width=0.77\columnwidth]{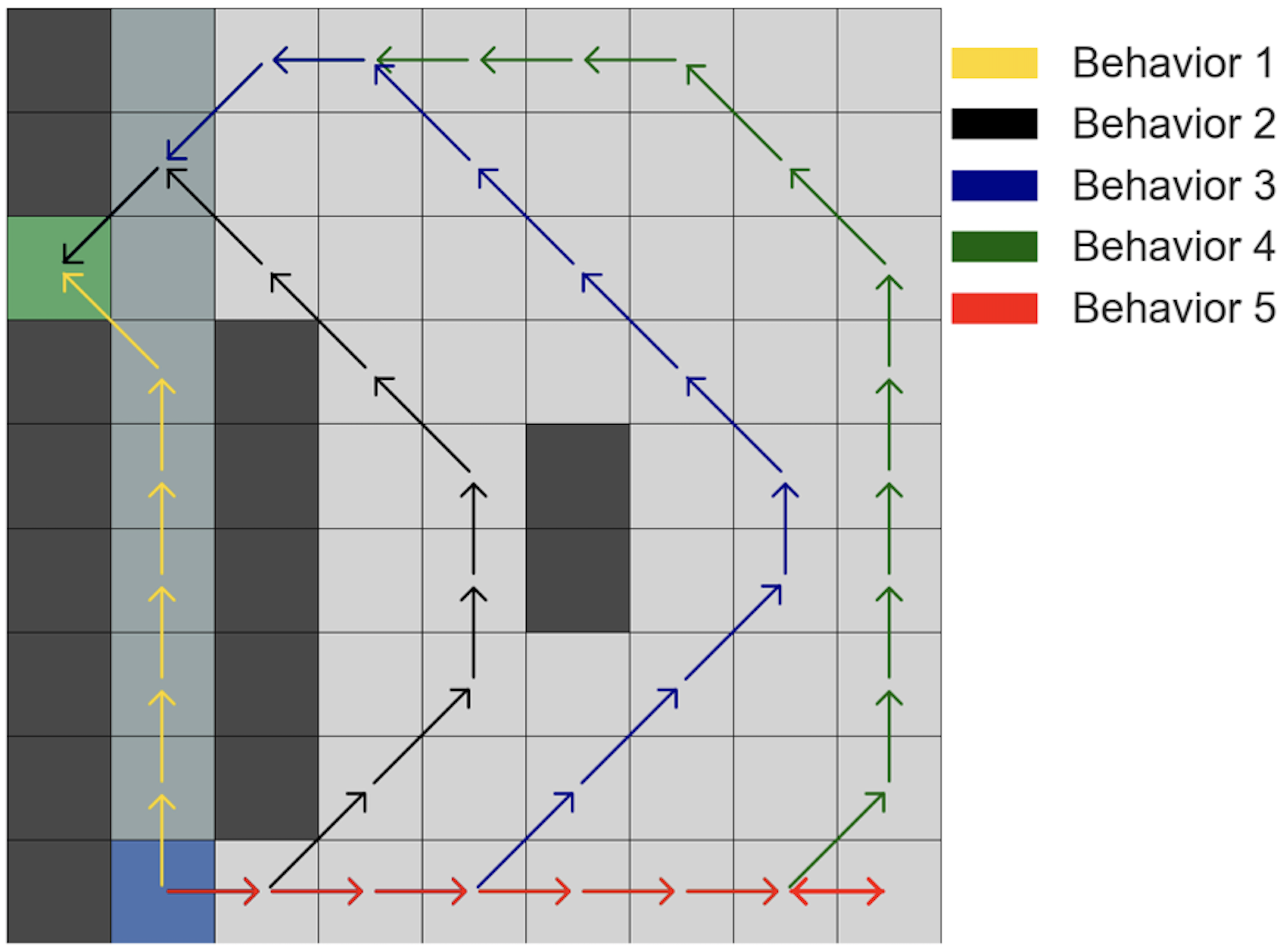}
}
\caption{(a) Grid World layout showing the reward structure. (b) {The five behavior profiles of risk-sensitive policies through the
        Grid World. These five paths correspond to the maximum likelihood paths of
        agents with various parameter combinations for their prospect,
        {\lprospect} and entropic map value functions. To generate each
        behavior with the prospect and {\lprospect} value functions, the following
        parameter combinations ($\{\kl,\ku,\Ll,\Lu\}$) were used: 
        \emph{Behavior 1}: $\{0.1,1.0,0.5,1.5\}$; \emph{Behavior 2}: $\{1.0,1.0,1.0,1.0\}$; \emph{Behavior
        3}: $\{1.0,1.0,1.1,0.9\}$; \emph{Behavior 4}: $\{5.0,1.0,1.1,0.8\}$; \emph{Behavior 5}: $\{5.0,1.0,1.5,0.7\}$.
    To generate the behaviors with the entropic map value function, we varied
    $\lambda$ from $1$ to $-1$.
}}
   \end{figure*}
\fi


\subsubsection{Setup}
Our instantiation of Grid World is shown in Fig.~\ref{fig:gridsetup}. An agent
operating in this MDP starts in the blue box and aims to maximize their value
function over an infinite time horizon. Every square in the grid represents a
state, and the action space is $A=\{N,NE,E,SE,$ $S,SW,W,NW\}$. Each action
corresponds to a movement in the specified direction (where we have used the
usual abbreviations for directions). The black and green states are absorbing,
meaning that once an agent enters that state they can never leave no matter
their action. In all the other states, the agent moves in their desired
direction with probability $0.93$ and they move in any of the other seven
directions with probability $0.01$. To make the grid finite, any action taking the agent out of the grid has
probability zero, and the other actions are re-weighted accordingly. The reward
structure of our instantiation of the Grid World is shown in
Fig.~\ref{fig:gridsetup} as well. The agent gets a reward of $-1$ and $+1$ for
being in the black and green states respectively. In the darker gray states, the agent gets a reward of $-0.1$. In all other states the agent is given a reward of $+0.1$.

\ifsc
\begin{table*}
\centering
\subfloat[ ][Learning with the Correct Model]{\begin{tabular}{|c ||r| r|| r| r||r|r|}
\hline
\multicolumn{1}{ |c|| }{\textbf{Value Function}} & \multicolumn{2}{ c||
}{\textbf{Prospect}} &  \multicolumn{2}{ c||}{\textbf{\lprospect}} 
& \multicolumn{2}{ c|
}{\textbf{Entropic}}\\
\cline{1-7}
\multicolumn{1}{|c||}{\textbf{\emph{Behavior}}} & \multicolumn{1}{c|}{\bf Mean}
& \multicolumn{1}{c||}{\bf Variance} & \multicolumn{1}{c|}{\bf Mean} &
\multicolumn{1}{c||}{\bf Variance} 
& \multicolumn{1}{c|}{\bf Mean} &
\multicolumn{1}{c|}{\bf Variance}\\
\hline\hline
\emph{Behavior 1}&$1.9$e-2   &$6.3$e-$4$ & $1.3$e-2 & $2.3$e-$4$ 
& $1.6$e-$3$ 
& $5.1$e-$6$\\\hline
\emph{Behavior 2} &$1.5$e-2   &$2.0$e-$4$ & $1.0$e-2 & $9.6$e-$5$ 
& $2.6$e-$4$ 
& $1.4$e-$7$\\\hline
\emph{Behavior 3} &$2.0$e-2   &$3.6$e-$4$ & $1.1$e-2 & $1.3$e-$4$ 
& $2.2$e-$3$ 
& $1.5$e-$5$\\\hline
\emph{Behavior 4} &$1.6$e-2   &$2.0$e-$4$ & $1.2$e-2 & $1.4$e-$4$ 
& $4.6$e-$4$ 
& $1.8$e-$7$\\\hline
\emph{Behavior 5} &$4.7$e-2   &$3.0$e-$3$ & $1.0$e-2 & $3.4$e-$4$ 
& $6.6$e-$4$ 
& $2.2$e-$7$\\
 \hline
\end{tabular}
\label{table:Grid}}\hspace*{0.2in} 

\subfloat[ ][Learning with an Incorrect Model]{\begin{tabular}{|c ||r| r|}
\hline
\multicolumn{1}{ |c|| }{\textbf{Value Function}} &\multicolumn{1}{ c| }{\bf Mean} &\multicolumn{1}{ c|
}{\bf Variance}\\
\hline\hline
Prospect       & $1.5$e-2    &$1.6$e-$4$ \\\hline
\lprospect     & $1.5$e-2    &$1.6$e-$4$ \\\hline
Entropic Map   & $5.4$e-2    &$1.4$e-$2$ \\
 \hline
\end{tabular}
\label{table:GridCross}}

\caption{{Mean and variance of the TV distance across all states in the grid of the
between the true policy and the policy under the learned value
function. We note that we present the best of five randomly sampled initial sets
of parameters. (a) Results for learning with the same type of
    value function as that of the agent. 
    (b) Results learning with different models than the true agent: 10,000
    trajectories are sampled from the policy of an agent with the prospect
    value function with $\{\kl,\ku,\Ll,\Lu\}=\{2.0,1.0,0.9,0.7\}$; prospect, {\lprospect}, and entropic map value
functions are learned from this data.}}
\end{table*} 

\else
\begin{table*}
\centering
\subfloat[ ][Learning with the Correct Model]{\begin{tabular}{|c ||r| r|| r| r||r|r|}
\hline
\multicolumn{1}{ |c|| }{\textbf{Value Function}} & \multicolumn{2}{ c||
}{\textbf{Prospect}} &  \multicolumn{2}{ c||}{\textbf{\lprospect}} 
& \multicolumn{2}{ c|
}{\textbf{Entropic}}\\
\cline{1-7}
\multicolumn{1}{|c||}{\textbf{\emph{Behavior}}} & \multicolumn{1}{c|}{\bf Mean}
& \multicolumn{1}{c||}{\bf Variance} & \multicolumn{1}{c|}{\bf Mean} &
\multicolumn{1}{c||}{\bf Variance} 
& \multicolumn{1}{c|}{\bf Mean} &
\multicolumn{1}{c|}{\bf Variance}\\
\hline\hline
\emph{Behavior 1}&$1.9$e-2   &$6.3$e-$4$ & $1.3$e-2 & $2.3$e-$4$ 
& $1.6$e-$3$ 
& $5.1$e-$6$\\\hline
\emph{Behavior 2} &$1.5$e-2   &$2.0$e-$4$ & $1.0$e-2 & $9.6$e-$5$ 
& $2.6$e-$4$ 
& $1.4$e-$7$\\\hline
\emph{Behavior 3} &$2.0$e-2   &$3.6$e-$4$ & $1.1$e-2 & $1.3$e-$4$ 
& $2.2$e-$3$ 
& $1.5$e-$5$\\\hline
\emph{Behavior 4} &$1.6$e-2   &$2.0$e-$4$ & $1.2$e-2 & $1.4$e-$4$ 
& $4.6$e-$4$ 
& $1.8$e-$7$\\\hline
\emph{Behavior 5} &$4.7$e-2   &$3.0$e-$3$ & $1.0$e-2 & $3.4$e-$4$ 
& $6.6$e-$4$ 
& $2.2$e-$7$\\
 \hline
\end{tabular}
\label{table:Grid}}\hspace*{0.2in} 
\subfloat[ ][Learning with an Incorrect Model]{\begin{tabular}{|c ||r| r|}
\hline
\multicolumn{1}{ |c|| }{\textbf{Value Function}} &\multicolumn{1}{ c| }{\bf Mean} &\multicolumn{1}{ c|
}{\bf Variance}\\
\hline\hline
Prospect       & $1.5$e-2    &$1.6$e-$4$ \\\hline
\lprospect     & $1.5$e-2    &$1.6$e-$4$ \\\hline
Entropic Map   & $5.4$e-2    &$1.4$e-$2$ \\
 \hline
\end{tabular}
\label{table:GridCross}}

\caption{{Mean and variance of the TV distance across all states in the grid of the
between the true policy and the policy under the learned value
function. We note that we present the best of five randomly sampled initial sets
of parameters. (a) Results for learning with the same type of
    value function as that of the agent. 
    (b) Results learning with different models than the true agent: 10,000
    trajectories are sampled from the policy of an agent with the prospect
    value function with $\{\kl,\ku,\Ll,\Lu\}=\{2.0,1.0,0.9,0.7\}$; prospect, {\lprospect}, and entropic map value
functions are learned from this data. }}
\end{table*}
\fi
\subsubsection{Learning with the correct model of the value function} 
This experiment is intended to validate our approach on a simple example.
We trained agents with various
parameter combinations of the four value functions described in
Section~\ref{sec:rsrl}. The resulting policies of these agents are classified
into five behavior profiles via
their maximum likelihood path through the MDP.
These behaviors are outlined in Fig.~\ref{fig:paths}. Each behavior
corresponds to the maximum likelihood path resulting from a different risk profile: \emph{Behavior 1} corresponds to 
a profile that is risk-seeking on gains, \emph{Behavior 2} corresponds to 
a profile that is risk neutral on gains and losses (this is also the behavior
corresponding to the non-risk-sensitive reinforcement learning approach), and 
\emph{Behaviors 3-5} correspond to behaviors that are increasingly risk averse on losses and increasingly weigh losses more than gains. 

We sampled 1,000 trajectories from the policies of these agents and used the
data to learn the value function of the agent using our gradient-based approach.
In this experiment, the learned value function is of the same type as that of
the agent. For example, the data sampled from the policy of an agent having a
prospect value function and exhibiting \emph{Behavior 1} is used to learn the
parameters of a prospect value function. We note that due to the non-convexity
of the problem, we use  five randomly generated initial parameter choices.

The results we report are associated with the value
function that achieves the minimum value of the objective. In
Table~\ref{table:Grid}, we report the mean TV distance between the two policies
across all states, as well as the variance in the TV distance across states.  
In all the cases considered in Table
\ref{table:Grid}, the learned value functions produce policies that correctly match the maximum
likelihood path of the {true} agent. 

We remark that the performance for
learning a prospect value function was consistently worse than learning an
{\lprospect} function. This is most likely due to the fact that the prospect value function is not Lipschitz around the reference point. Thus, we have no
guarantees of differentiability of $Q^\ast$ with respect to $\theta$ for the
prospect value function. This translates to numerical issues in calculating the
gradient which, in turn, results in worse performance.

The entropic value function performs best of the four value functions, primarily
due to the fact that there is only one parameter to learn, and the rewards and
losses are all relatively small. In fact, in all the cases the learned entropic
map value function coincided with the true value function of the agent, thereby indicating that the objective function was relatively convex around the parameter values we tried. 

%

\subsubsection{Learning with an incorrect model of the value function} 
The second experiment consists of learning different types of value functions
from the same dataset. This is a more realistic experiment since the value
function of human subjects will very likely be different than any model we could
choose. The motivation for this experiment is to ensure that the results and risk-profiles learned were consistent across our choice of model.

The experiment uses 10,000 samples from an agent with a prospect value function and
learned prospect, {\lprospect}, and entropic map value functions. The
mean TV distance between the policy of the {true} agent and the policies under
the learned value functions are shown in Table~\ref{table:GridCross}. The true
agent's value function has parameters
$\{\kl,\ku,\Ll,\Lu\}=\{2.0,1.0,0.9,0.7\}$---that is, it is risk-seeking in
losses, risk-averse in gains, and loss averse.

Again, the learned value functions all have policies that replicated the maximum
likelihood behavior of the true agent. We note that the {\lprospect} and prospect
functions perform as well as each other on this data, but the {\lprospect} function
showed none of the numerical issues that we encountered with the prospect
function (see
Section~\ref{sec:numerical} for further detail on numerical considerations).
Further, learning with the {\lprospect} function is markedly faster than with
the prospect function. Again, this is most likely due to the fact that the prospect
function is not locally Lipschitz continuous around the reference point. Thus,
the values of $\alpha$ required to make the various contraction maps converge to
their fixed points are vanishingly small. This results in slow convergence. 

The fact that the entropic value function does not perform
as well is most likely due to the fact that 
it cannot accurately match the shape of the prospect function at these values; e.g., the
entropic map is always either convex or concave.
\ifsc
\begin{figure}[]
    \centering
    \includegraphics[width=0.5\columnwidth]{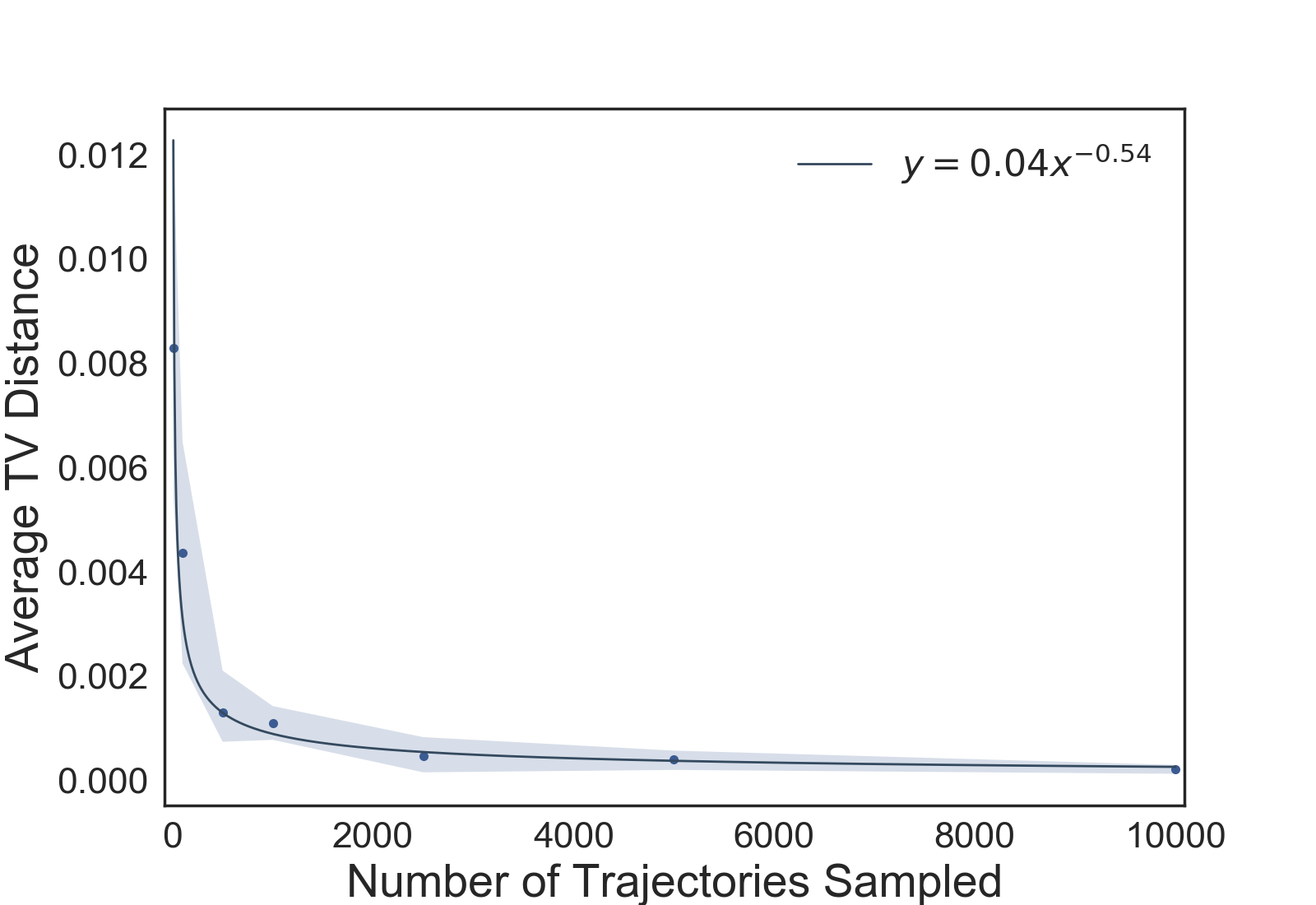}
    \caption{{
The mean TV distance across all states between the agent's policy and the
policy under the learned value function, as a function of the number of
trajectories in the dataset. To construct each data point, we sample five
different datasets of the same number of trajectories from the agent's policy.
We try five random initial parameter values 
per dataset and take the value function that achieves the minimum value
of the objective. We calculate the mean TV distance between the policy
of the agent and the policy under the learned value function for each dataset
and then average these values. The bars show the 95\% confidence interval
around the mean of the five datasets of the given size. We note that the trendline $y=0.04x^{-0.54}$ is the best fit of the form $y=ax^b$ to the data points, for constant terms $a,b$. } }
    \label{fig:samplecomplexity}
\end{figure}

\else
\begin{figure}[]
    \centering
    \includegraphics[width=0.7\columnwidth]{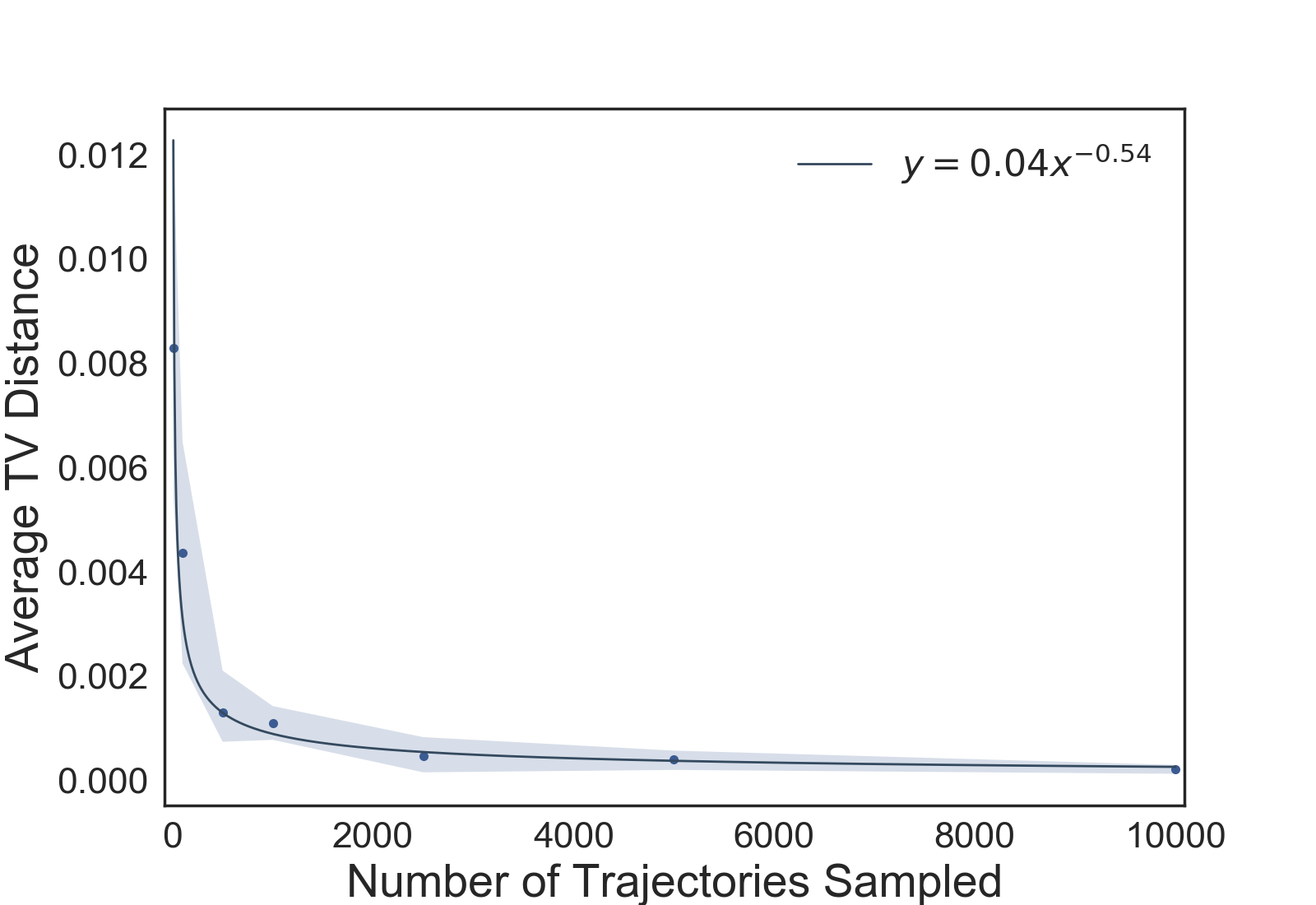}
    \caption{{
The mean TV distance across all states between the agent's policy and the
policy under the learned value function, as a function of the number of
trajectories in the dataset. To construct each data point, we sample five
different datasets of the same number of trajectories from the agent's policy.
We try five random initial parameter values 
per dataset and take the value function that achieves the minimum value
of the objective. We calculate the mean TV distance between the policy
of the agent and the policy under the learned value function for each dataset
and then average these values. The bars show the 95\% confidence interval
around the mean of the five datasets of the given size. We note that the trendline $y=0.04x^{-0.54}$ is the best fit of the form $y=ax^b$ to the data points, for constant terms $a,b$. } }
    \label{fig:samplecomplexity}
\end{figure}
 \fi
\subsubsection{Qualitative results on sample complexity}
\label{sec:results_complex}
One of the challenges in modeling human decision-making is the lack of access to
large datasets, particularly when it comes to sequential decisions that are made
over longer periods of time.   This is counter to the usual learning scenarios addressed in the much of the learning
literature. For instance, if the focus is learning to control a robot, then it
may be possible to generate a large number of demonstrations very quickly. This
motivates our third experiment with the Grid World MDP---i.e.~an experiment
that allows us to better
understand how the performance of our approach varies with the size of the
dataset. 

In this experiment, we first train an agent with an entropic map value
function and then create sets of sample
 trajectories from the agent's policy varying between zero and 10,000 in size.
Next, using each of these sample sets, we learn the
value function via our approach and plot the mean TV distance across all states between the
true policy of the agent and the policy under the learned value function. This
is shown in Fig.~\ref{fig:samplecomplexity}.  

First, we note that more data does translate to consistently better results.
This matches our intuition that the better our data matches the policy of the
true agent, the better we can learn a value function that would be associated
with that policy. Of particular interest, though, is the rate at which the
average TV across all the states decreases with the number of trajectories
sampled. The rate, which is on the order of $x^{-0.54}$, is very close to the asymptotic
rate, derived in Section~\ref{sec:complexity}, of $O(x^{-1/2})$. This suggests
that the dominating factor in the performance of our algorithm is how well our
data matches the underlying policy, and not the non-convexity of the objective
function. In fact, this provides empirical evidence that the second term in
\eqref{eq:bnd}---i.e.~$\delta(\hat{\pi}_n(\cdot|x),\pi_\theta(\cdot|x))$---must
also be $O(x^{-1/2})$.
\subsection{A Passenger's View of Ride-Sharing}

In addition to the Grid World example, we explore a ride-sharing example for
which the MDP is created from real-world data and we simulate agents with
different risk preference and loss aversion profiles\footnote{We adopt the
    \emph{surge pricing} model here due to the availability of
data even though ride-sharing
    services such as Uber are moving
    towards personalized pricing schemes that offer prices that the rider is
    \emph{willing to pay}.
    This kind of pricing model motivates even more strongly the
    need for techniques that are considerate of how humans actually make
decisions.}.




Many ride-sharing companies set prices based on both supply of drivers and demand of
passengers.
From the passenger's viewpoint, we model the ride-sharing MDP as follows. The
action space is $A=\{0,1\}$ where $0$ corresponds to `wait' and $1$
corresponds to `ride.' The state
space $X=\mc{X}\times \mc{T}\cup \{x_{\text{f}}\}$ where $\mc{X}$ is a finite
set of surge price multipliers,
$\mc{T}=\{0, \ldots,
    T_{\text{f}}\}$ is the part of the state corresponding to the time
    index, and $x_{\text{f}}$ is a terminal state representing the completed
    ride that
    occurs when a ride is taken. At time $t$, the state is notationally given by
    $(x_t,t)$. The reward $r_t$ is modeled as a random variable that depends on the
    current price as well as a random variable $Z(t)$ for travel time.
    In particular, for any time $t<T_{\text{f}}$ the reward is given by
    \begin{equation}
        R(x_{t},a_t)=\left\{\begin{array}{ll}
            \bar{r}, &\   a_t=0 \ (\text{'wait'})\\
            \tilde{r}_t, & \ a_t=1\ (\text{'ride'})\end{array}\right.
        \label{eq:R}
    \end{equation}
    with $\bar{r}<0$ a constant and 
    $\tilde{r}_t={S}_t-x_t(p_{\text{base}}+p_{\text{mile}}D+p_{\text{min}}Z(t))$ 
    where $D$ is the distance in miles, ${S}_t$ is a time dependent
    satisfaction (we selected it to linearly decrease in time from some initial
    satisfaction level), and $p_{\text{base}}$, $p_{\text{mile}}$, and
    $p_{\text{min}}$ are the base, per mile, and per min prices,
    respectively. 
    
    At the final time $T_{\text{final}}$, the agent is forced to take the ride  if
    they have not selected to take a ride at a prior time. This reflects the
    fact that the agent presumably needs to get from their origin to their
    destination and the reward structure
    reflects the dissatisfaction the agent feels as a result of having to
    ultimately take
    the ride despite the potential desire to wait.

         Using the Uber Movement\footnote{Uber Movement: {\tt
  \url{https://movement.uber.com/cities}}}
            platform for travel time statistics, base pricing data\footnote{The base, per min, and per mile prices can be found here:
        {\tt \url{http://uberestimate.com/prices/Washington-DC/}}} and surge
        pricing data\footnote{The surge pricing data we used was originally collected by and has
been made publicly available here: {\tt
    \url{https://github.com/comp-journalism/2016-03-wapo-uber}}. The data we use was
collected over three minute intervals in period between November 14 to November 28, 2016.} for
        Washington D.C., we examined several locations and hours which have different
          characteristics in terms of travel time and price statistics. We
          generate the distribution for $Z(t)$ from these data sets as well as the
          surge price intervals and transition probability matrix. Since
          the core risk-sensitive behaviors we observe are similar across the
          different locations, we report only on one.
        
            Specifically, we report on  a ride-sharing MDP generated with an 
          origin and destination of GPS$=(-77.027046,$ $38.926749)$ and GPS$=(-76.935773,$ $38.885964)$\footnote{Note that these correspond to Uber Movement id's 197
and 113, respectively.}, respectively, in Washington
D.C.~at 5AM.
\ifsc
\begin{table*}[t]
    \centering
\begin{tabular}{|c||l|l||l|l||l|l|}
  \hline
  \multicolumn{1}{ |c|| }{\textbf{{Value Function}}} &
  \multicolumn{2}{c||}{\textbf{{Prospect}}} &
  \multicolumn{2}{c||}{\textbf{{Entropic}}} &
  \multicolumn{2}{c|}{\textbf{{Prospect/\lprospect}}} \\
  \cline{1-7}
  \textbf{ \emph{Preferences}}& \textbf{Mean} & \textbf{Variance} &\textbf{Mean} &
  \textbf{Variance} & \textbf{Mean} & \textbf{Variance}   \\
  \hline    \hline
  \emph{RA Gains/RS Losses (Entropic: RA)}& 1.3e-2 & 3.5e-4 & 0.9e-3 & 1.0e-6 & 1.0e-2 & 1.4e-4  \\      \hline
  \emph{ Risk-Neutral    }    & 0.6e-2 & 4.9e-5 & 1.4e-3 & 2.0e-6 & 6.6e-3 & 1.0e-4  \\     \hline
  \emph{RS Gains/RA Losses (Entropic: RS)} & 1.1e-2 & 1.7e-4 & 1.1e-3 & 1.5e-6 & 1.1e-2 & 1.1e-4  \\     \hline

\end{tabular}
\caption{{Averaged TV error and variance over 10 different initializations of
    the algorithm for different risk-preference profiles.
    The last column shows
    the error when using an {\lprospect} agent with $\epsilon=1$e-2 to learn a
prospect agent. We use the following abbreviations:  Risk-Averse (RA);
Risk-Seeking (RS). }}
    \label{tab:ubererrors}
\end{table*}

\else
\begin{table*}[t]
    \centering
\begin{tabular}{|c||l|l||l|l||l|l|}
  \hline
  \multicolumn{1}{ |c|| }{\textbf{{Value Function}}} &
  \multicolumn{2}{c||}{\textbf{{Prospect}}} &
  \multicolumn{2}{c||}{\textbf{{Entropic}}} &
  \multicolumn{2}{c|}{\textbf{{Prospect/\lprospect}}} \\
  \cline{1-7}
  \textbf{ \emph{Preferences}}& \textbf{Mean} & \textbf{Variance} &\textbf{Mean} &
  \textbf{Variance} & \textbf{Mean} & \textbf{Variance}   \\
  \hline    \hline
  \emph{Risk-Averse Gains/Risk-Seeking Losses (Entropic: Risk-Averse)}& 1.3e-2 & 3.5e-4 & 0.9e-3 & 1.0e-6 & 1.0e-2 & 1.4e-4  \\      \hline
  \emph{ Risk-Neutral    }    & 0.6e-2 & 4.9e-5 & 1.4e-3 & 2.0e-6 & 6.6e-3 & 1.0e-4  \\     \hline
  \emph{Risk-Seeking Gains/Risk-Averse Losses (Entropic: Risk-Seeking)} & 1.1e-2 & 1.7e-4 & 1.1e-3 & 1.5e-6 & 1.1e-2 & 1.1e-4  \\     \hline

\end{tabular}
\caption{{Averaged TV error and variance over 10 different initializations of
    the algorithm for different risk-preference profiles.
    The last column shows
    the error when using an {\lprospect} agent with $\epsilon=1$e-2 to learn a prospect agent.}}
    \label{tab:ubererrors}
\end{table*}
\fi

\begin{figure*}[h!]
    \subfloat[][risk-seeking in gains/risk-averse in losses (convex/concave),
$\Lu=\Ll=1.5$]{
\includegraphics[width=0.3\textwidth]{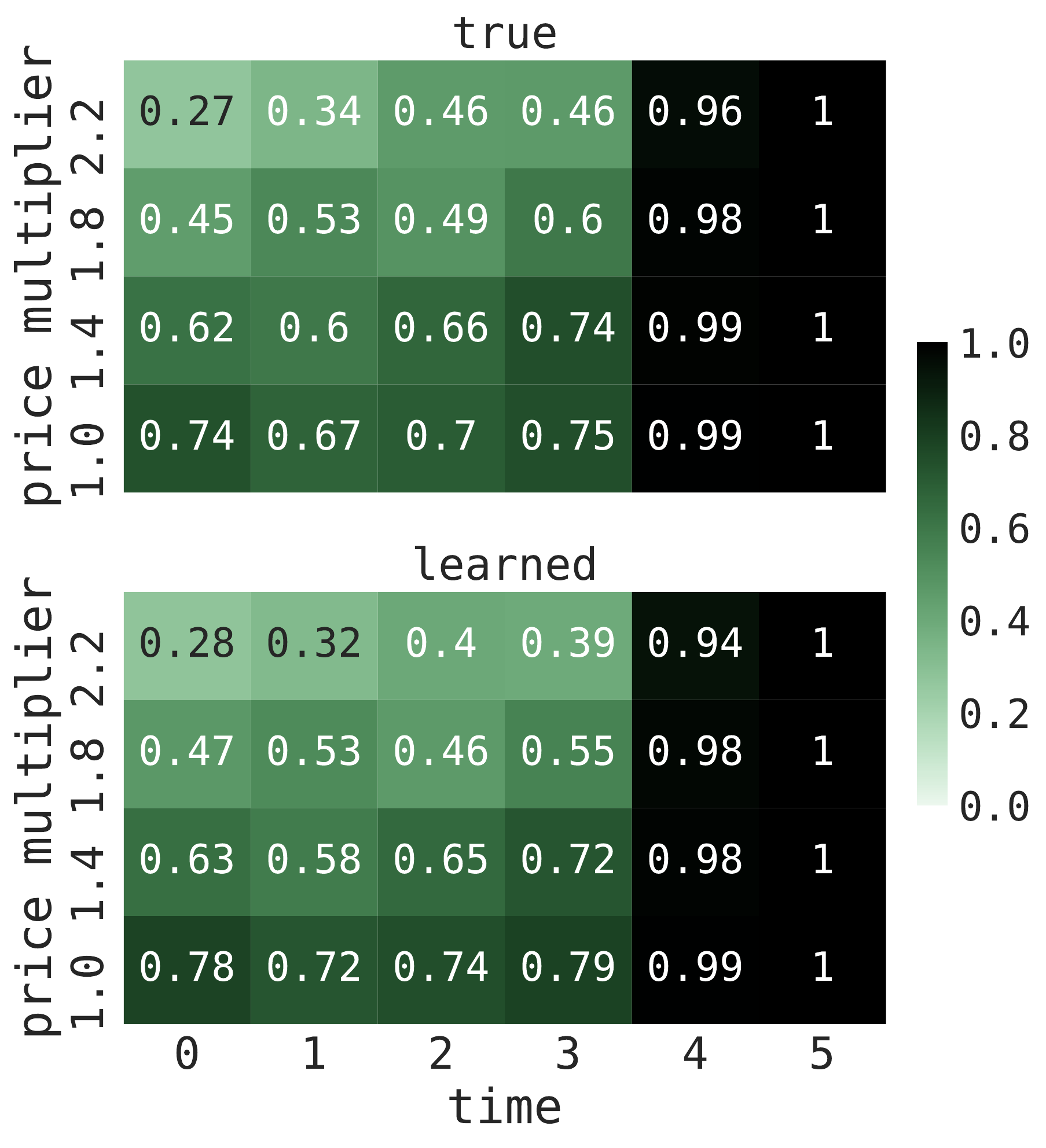}\label{fig:RSRA}} 
\hspace*{0.1in}
\subfloat[][risk-neutral,
$\Lu=\Ll=1.0$]{\includegraphics[width=0.3\textwidth]{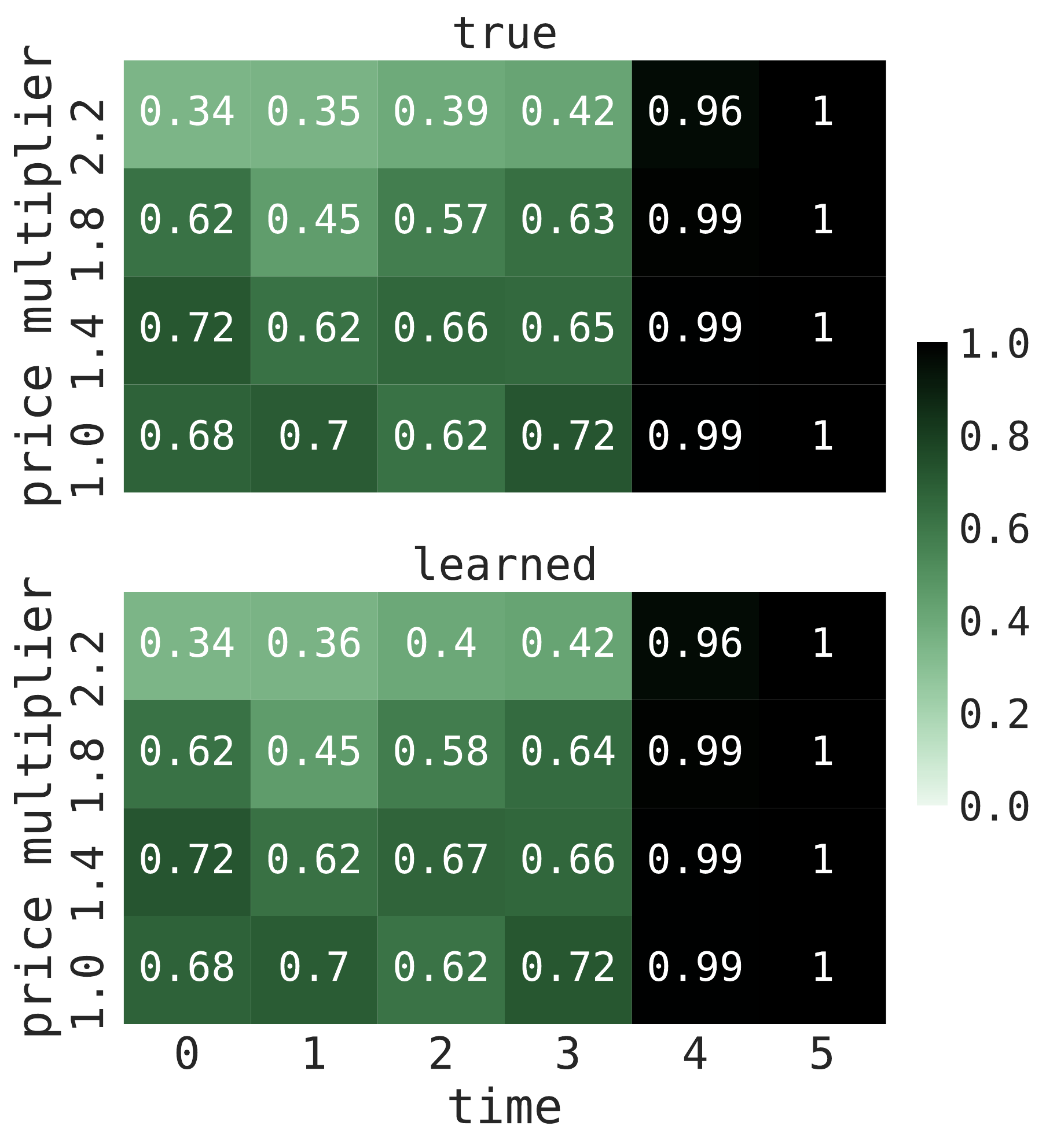}\label{fig:RN}}
\hspace*{0.1in}   
{\subfloat[][risk-averse in gains/risk-seeking in losses (concave/convex),
$\Lu=\Ll=0.5$]{ 
    \includegraphics[width=0.3\textwidth]{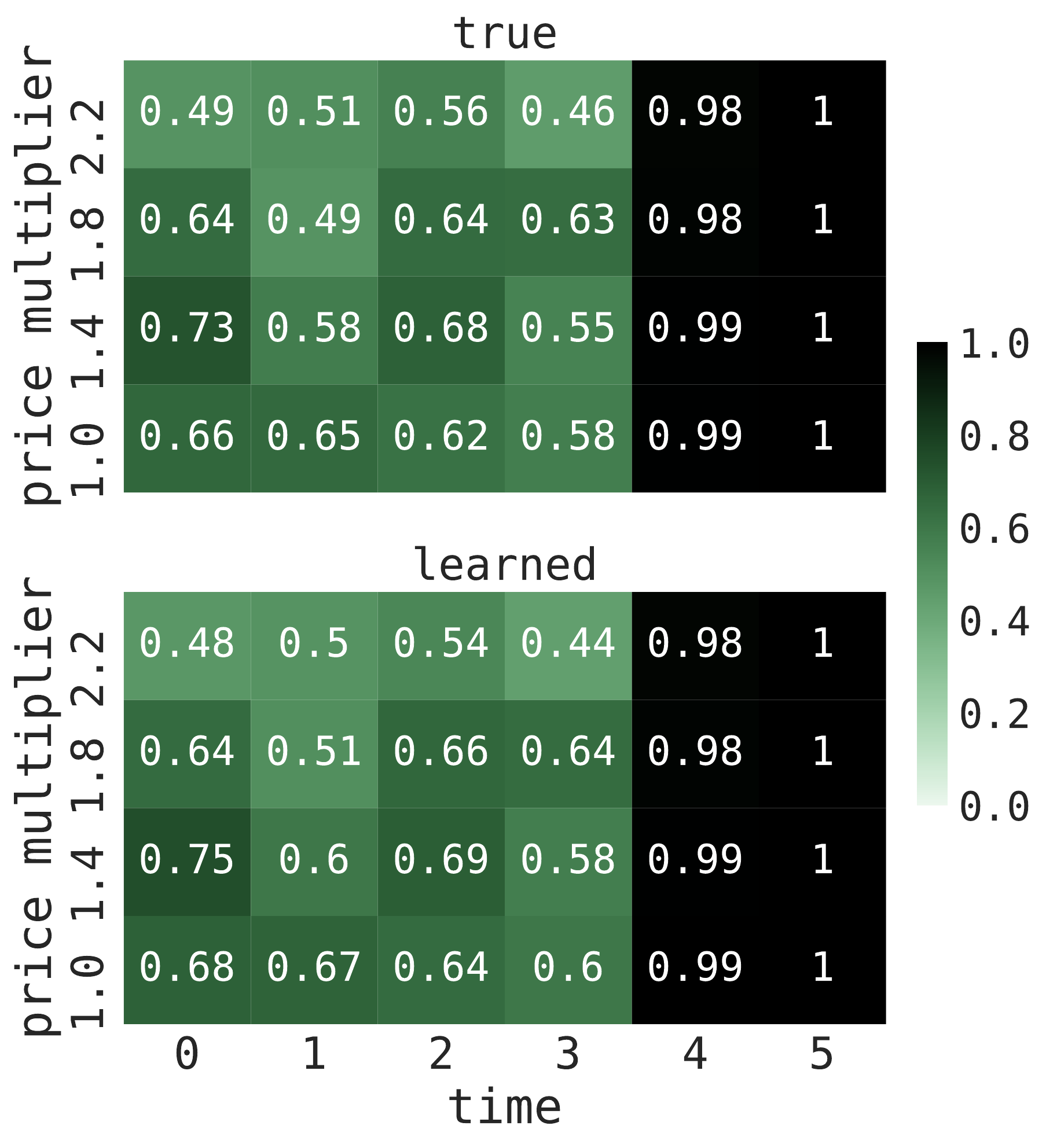}\label{fig:RARS}}}
\caption{
Plots showing the probabilities of taking a ride in each state
under the true and learned optimal policies for true and learned
agents with prospect value functions. The true agent has prospect
gain parameters of $\ku=0.5$ and $\kl=1.0$ for all three plots. The value function used for the right most graphic
(Fig.~\ref{fig:RARS}) is most representative of human decision-making since
humans tend to be risk-averse in gains, risk-seeking in losses, are loss averse.
In these plots, the trend we
see is that the more risk-averse, the less likely they policy suggests taking the ride.}
\label{fig:uber-new}
\end{figure*}

The transition probability kernel $P:X\times A\times X
    \rar [0,1]$ is estimated from the ride-sharing data. The  travel-time data 
    is available on an hourly basis and the price change data is available on a
    three minute basis. Hence, we use the three minute price change
    data for each hour to derive a static transition
    matrix by empirically estimating the transition probabilities where we bin
    prices in the following way. For prices in $[1.0,1.2)$, $x=1.0$;
        for prices in $[1.2,1.6)$, $x=1.4$; for prices in $[1.6,2.0)$, $x=
            1.8$; otherwise $x=2.2$. Hence, $\mc{X}=\{1.0,1.4,1.8,2.2\}$. In the time periods we examine,
            the max price multiplier was $2.2$. We set the reference point
            $y_0$ and
            acceptance level $v_0$ to be zero.

With this model, the transition matrix for the price multipliers is given by
\begin{equation}
  P=  \bmat{ 0.876 &  0.099 & 0.017 &  0.008\\
       0.347&  0.412&  0.167 & 0.074\\
       0.106 &  0.353 &  0.259 & 0.282\\
   0.086 &  0.219 &  0.143 &  0.552}
    \label{eq:transmat}
\end{equation}
for each time.
The travel time distribution is a standard normal distribution truncated to the
upper and lower bounds specified by the Uber Movement data. Measured in seconds,
we use location
parameter $2371$, scale parameter $100$, and $1554$ and
$3619$ as the upper and lower bound, respectively.

\begin{figure*}[h!]
    \subfloat[][less loss-averse
$\ku=1.5, \kl=0.5$]{
\includegraphics[width=0.3\textwidth]{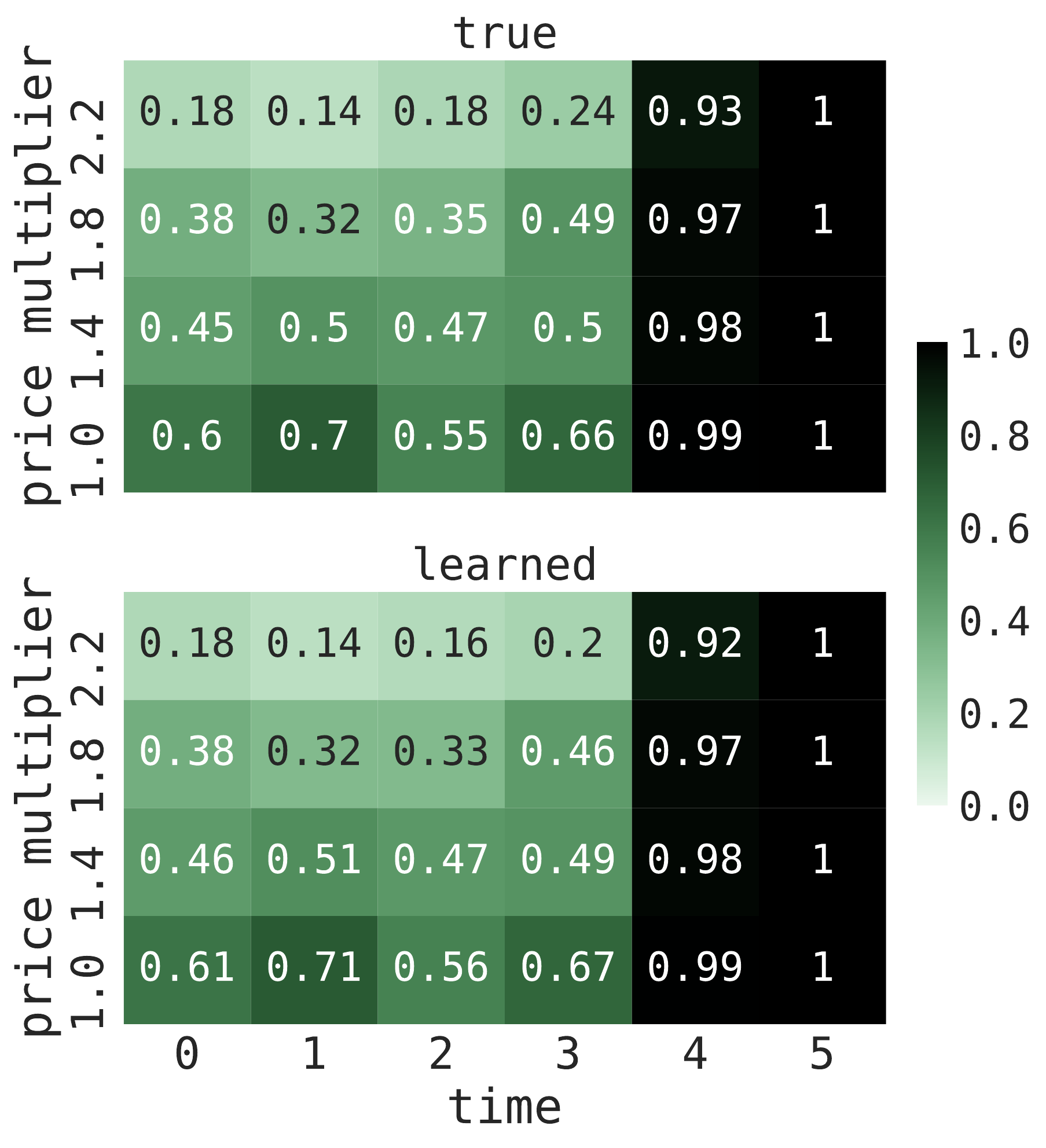}\label{fig:RSRA}} 
\hspace*{0.1in}
\subfloat[][no loss-aversion,
$\ku=\kl=1.0$]{\includegraphics[width=0.3\textwidth]{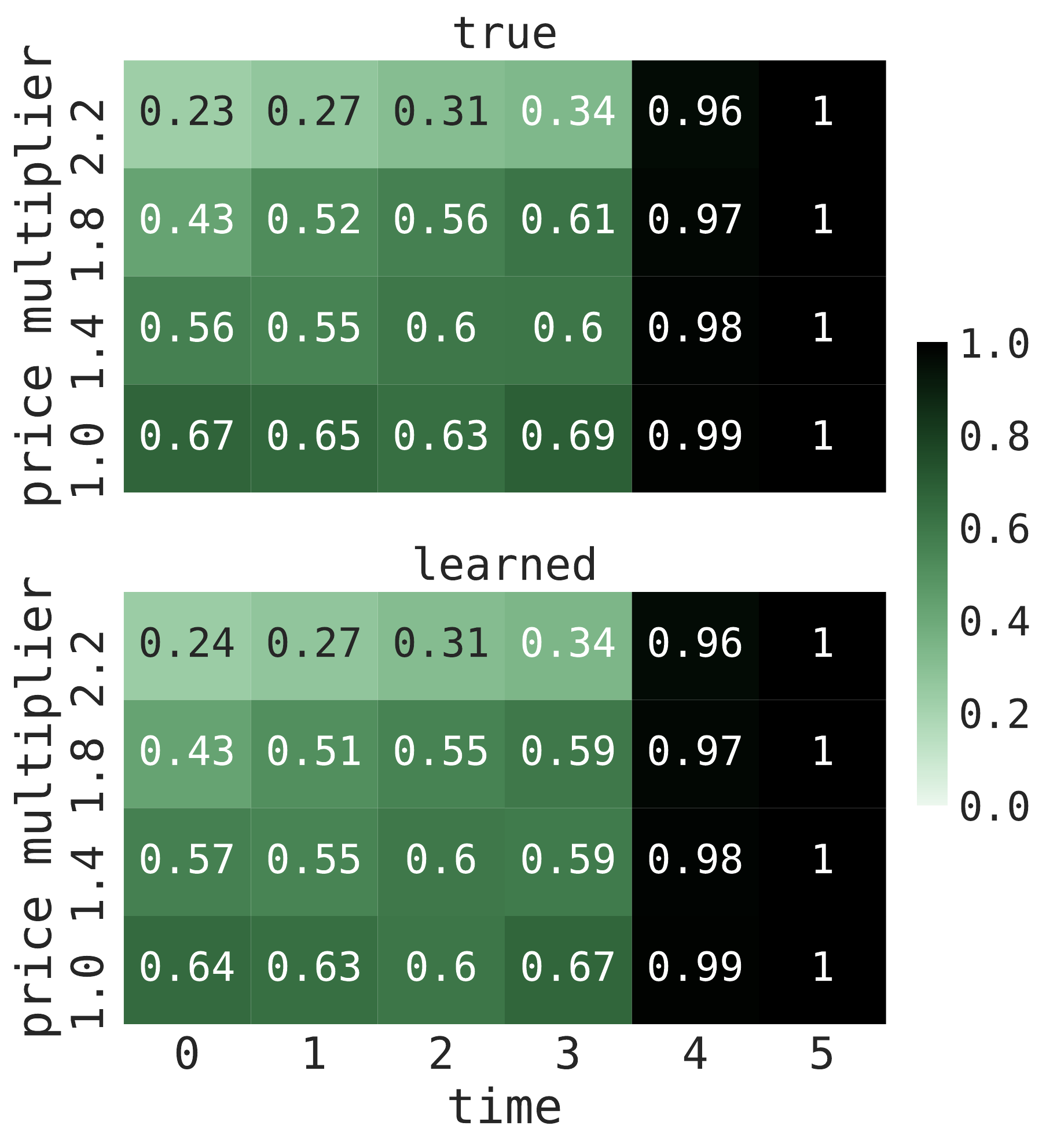}\label{fig:RN}}
\hspace*{0.1in}   
{\subfloat[][more loss-averse,
$\ku=0.5, \kl=1.5$]{ 
    \includegraphics[width=0.3\textwidth]{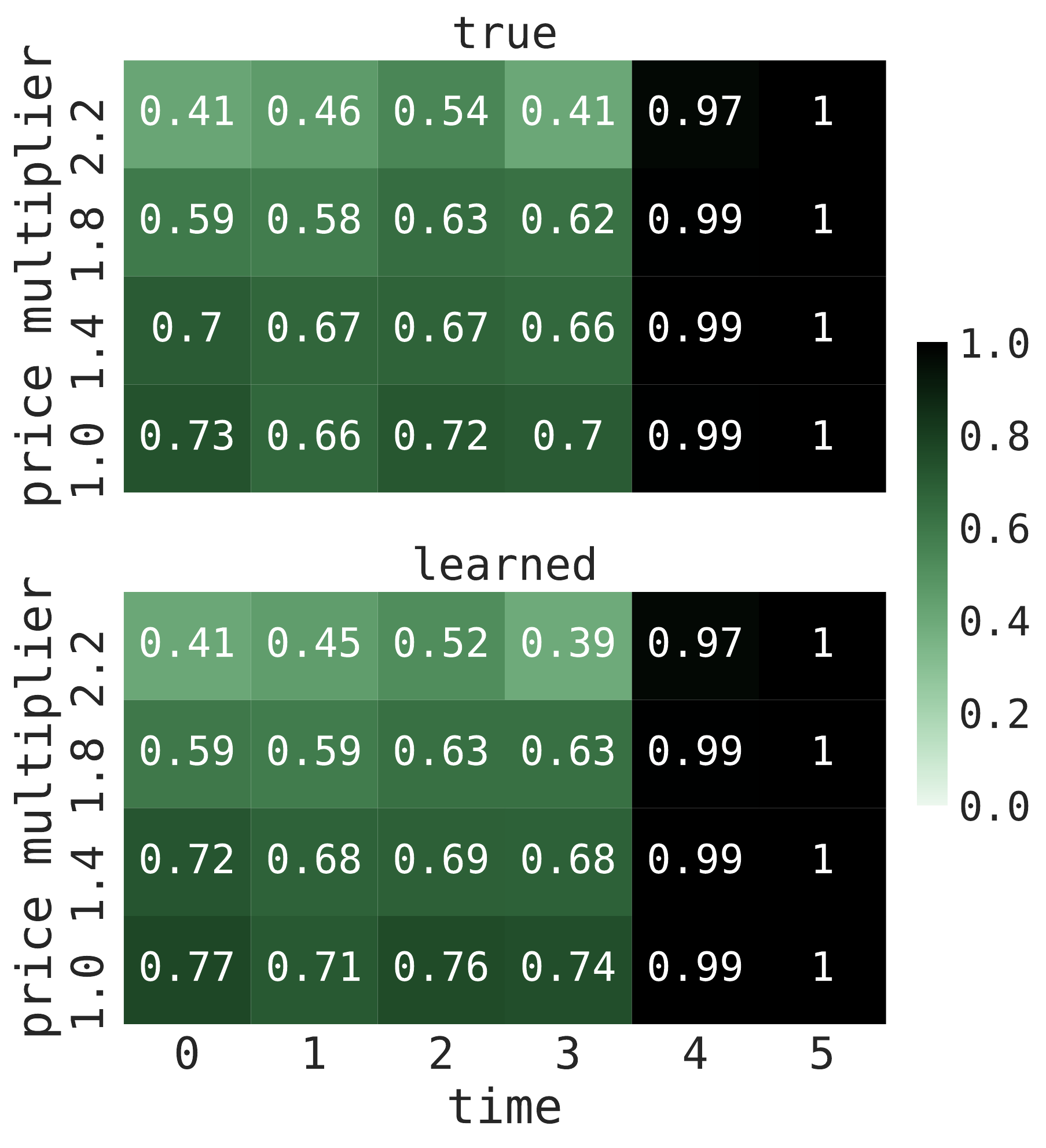}\label{fig:RARS}}}
\caption{
Plots showing the probabilities of taking a ride in each state
under the true and learned optimal policies for true and learned
agents with prospect value functions. The true agent has prospect
parameters of $\Ll=\Lu=1.0$ for all three plots, while we vary $(\ku,\kl)$ to
capture different degrees of loss-aversion. 
In these plots, the trend we
see is that the more loss-averse the agent (under both the learned and true
value functions), the more likely they are to take the ride.}
\label{fig:uber-new2}
\end{figure*}
\ifsc
\begin{figure}[t]
    \centering
\includegraphics[width=0.4\columnwidth]{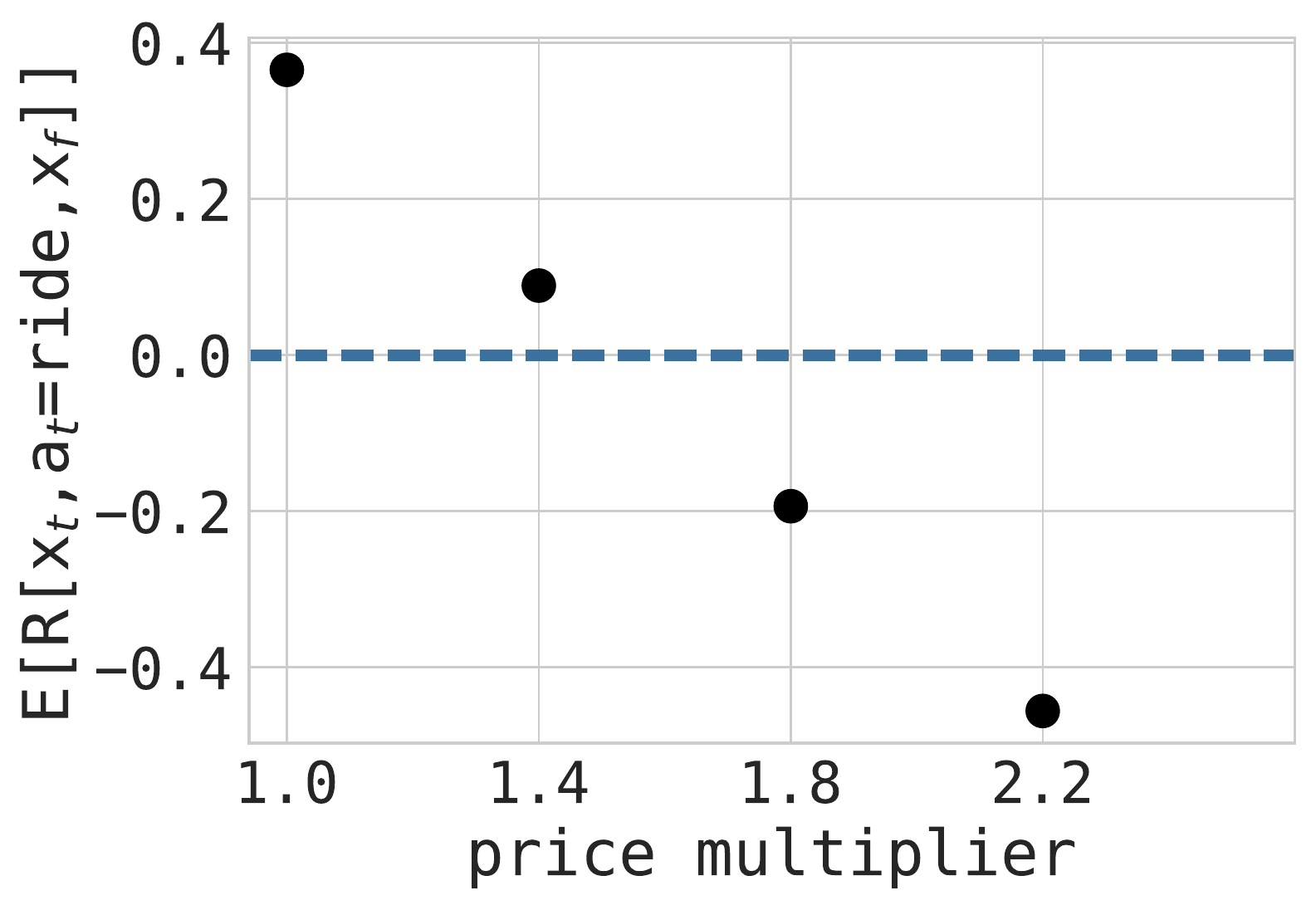}
        \caption{Expected rewards for each time step in the ride-sharing example. Notice that the rewards
        can either be gains (positive values) or losses (negative values) given that we take the reference point to
    be $y_o=0$. }
    \label{fig:rewards}
\end{figure}
\else
\begin{figure}[t]
    \centering
\includegraphics[width=0.6\columnwidth]{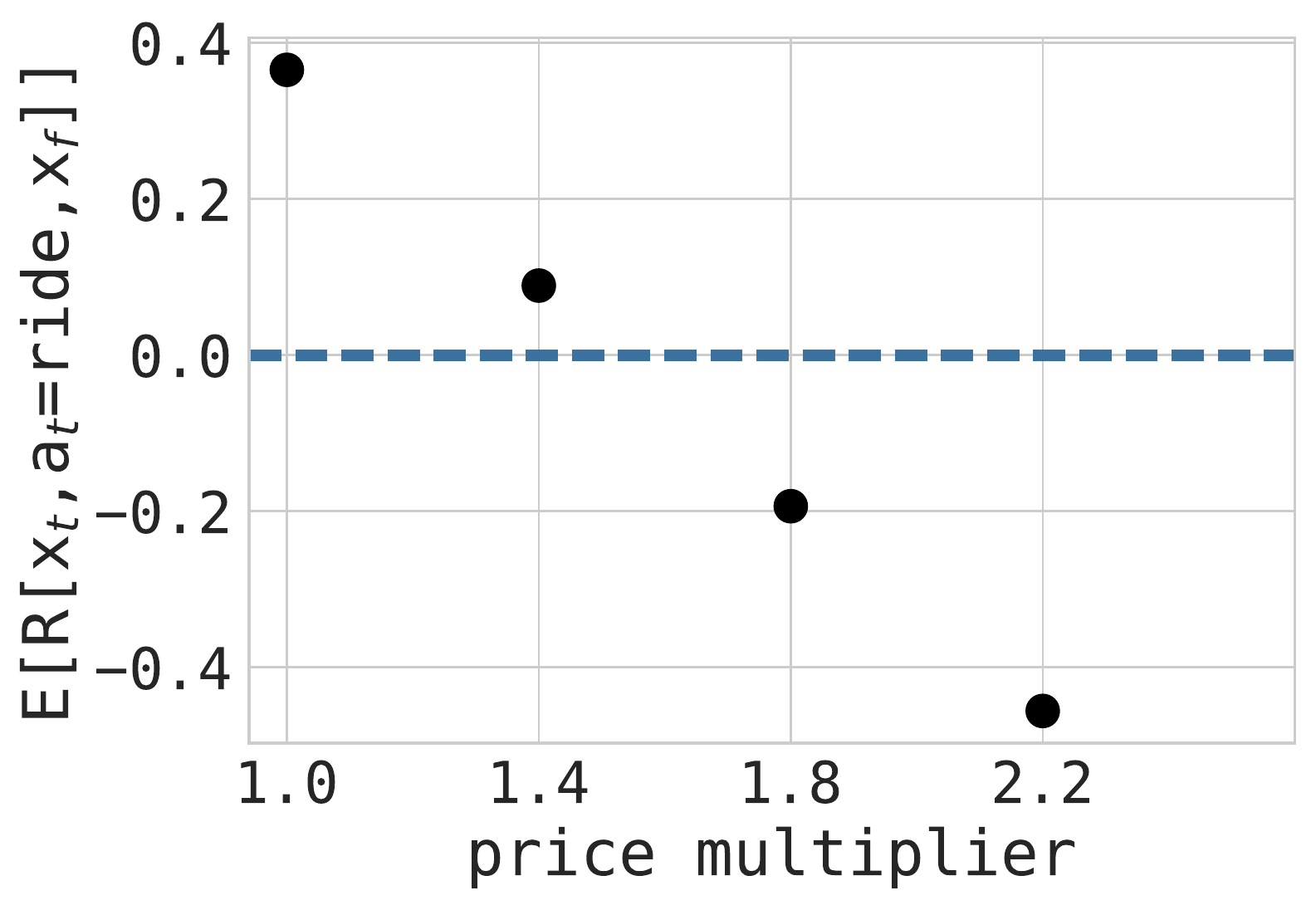}
        \caption{Expected rewards for each time step in the ride-sharing example. Notice that the rewards
        can either be gains (positive values) or losses (negative values) given that we take the reference point to
    be $y_o=0$. }
    \label{fig:rewards}
\end{figure}
\fi
        
The graphics in Fig.~\ref{fig:uber-new} and Fig.~\ref{fig:uber-new2} show the state space as a grid with the probability of taking a ride under the
true and learned optimal policies overlaid on each state.
For the examples depicted in these figures, we consider the true and learned
agents to have 
prospect value functions.  

In Fig.~\ref{fig:uber-new}, we fix the true agent's parameters to show a range of behaviors from
risk-seeking in losses/risk-averse in gains to risk-averse in
losses/risk-seeking in gains.  There is empirical evidence supporting the fact
that humans are more like the former. Moreover, in these examples we use
$(\ku,\kl)=(0.5,1)$ to capture that humans tend to be loss-averse---that is, for losses
and gains of equal value, the loss is perceived as more significant.

On the other hand, in Fig.~\ref{fig:uber-new2} we fix the true agent's
parameters to show a range of behaviors depending on the degree of
loss-aversion. In particular, we fix $\Ll=\Lu=1$ and vary the ratio of $\kl$
to $\ku$, where a higher ratio corresponds to more loss averse preferences.

In each of the graphics in Fig.~\ref{fig:uber-new} and Fig.~\ref{fig:uber-new2}, we see that the learned policy is very close to the
true policy. In addition, in Fig.~\ref{fig:uber-new}, we observe that the more risk-averse the agent is (in
gains or losses), the more likely they are to take the ride. This trend can be seen by
noting the sign of the expected rewards---in Fig.~\ref{fig:rewards}, we see that
the reward is
positive for $x_t\in\{1.0,1.4\}$ and is negative for $x_t\in\{1.8,2.2\}$---and
examining the
corresponding rows in Fig.~\ref{fig:uber-new} for negative and positive rewards.
In Fig.~\ref{fig:uber-new2}, observe that the more loss averse the
agent, the
more likely they are to take the ride uniformly. This is reasonable as the
satisfaction level is linear decreasing in time.

In Table~\ref{tab:ubererrors}, we show the mean and variance of the total
variation error for the ride-sharing example where we varied the
risk preference profiles, holding $(\kl,\ku)=(1,1)$, using agents with prospect and
entropic value functions. In addition, we show the error for different risk
profiles when we learn a true prospect agent with an {\lprospect} agent. Recall
that the prospect value function does not meet the requirements of our theorem
whereas the {\lprospect} value function does as it is Lipschitz. 


\subsection{Numerical Considerations}
\label{sec:numerical}

We end the experimental results section with some observations on the convergence speed 
and the implementation of Algorithm 1. 

First, we note that  the 
two contraction mappings \eqref{eq:Qtup-2} and
\eqref{eq:pdpi} are sensitive to the learning rate $\alpha$. 
A very small choice of $\alpha$
results in convergence of the sequence of Q-functions to the fixed point being
too slow to be practically useful. On the other hand, a large choice of $\alpha$
makes the sequence diverge. Thus, choosing $\alpha$ has a large effect on the
runtime of the overall algorithm as the computation of $Q^\ast$ and
$D_\theta Q^\ast$
both depend on the choice of $\alpha$. 

We further remark that  numerical observations suggest that the condition
$\alpha \in (0,\min\{L^{-1},1\}]$ is
fairly restrictive and that larger values of $\alpha$ give faster convergence.
Hence, our implementation of Algorithm 1 includes an adaptive scheme to find the largest possible $\alpha$. 
In particular, if two consecutive iteration elements in the sequence are
observed to diverge in the $L_\infty$ norm, we decrease $\alpha$ by a fixed
constant. As long successive elements in the sequence converge, we periodically
increase $\alpha$ by another constant. This allows us to noticeably speed up the
implementation of our algorithm. Adaptively choosing the step-size $\alpha$ also
allows us to train the prospect function agents more accurately, since these were particularly susceptible to changes in the value of $\alpha$ due to the fact that the value function is non-Lipschitz around the reference point.

To speed up the gradient-descent algorithm, we also implement a back-tracking
line search. We do this to address the computationally intensive gradient
calculation.
Specifically, the line-search allows us to exploit each gradient calculation
fully. The backtracking line search also leads to a noticeable speed up in the
implementation of our algorithm, which allows us to tackle larger MDPs.

\section{Related Work}
\label{sec:relatedwork}
Before concluding, we draw some comparisons between our work and that of others
on related topics. 

The primary motivation for most other works in this domain is to
        learn a prescriptive model  for humans amidst autonomy.
        For example,
        in~\cite{shen:2014aa},
        their 
         approach to learning the decision-making model is to parameterize
        unknown quantities of interest, sample the parameter space, and use a
        model selection criteria (i.e.~Bayesian information
        criteria) to select parameters that best fit the observed behavior. In
        contrast, we
       derive a well-formulated gradient-based procedure
        for finding the value function and policy best matching the
        observed behavior. Moreover, we introduce new value functions that
        satisfy our theorems for the forward and inverse problems while retain the salient features of the empirically
        observed behavioral models.

        Another related work is~\cite{majumdar:2017aa}.  The authors
        take a similar approach to ours in leveraging risk metrics to
        capture risk sensitivity. However, they focus their efforts on estimating the
        risk metric  by leveraging the well-known
        representation theorem for coherent risk metrics~\cite{follmer:2002aa}.
    They couple the resulting   optimization problem with classical inverse reinforcement learning
        procedures for learning the reward (that is, they parameterize the
        reward function over a set of basis functions), yet their approach does not
        differentiate between the reward and the decision-making model.
 In contrast,
 we consider a broad class of risk metrics generated by
 value functions via acceptance sets, formulate the MDP model based on the risk
 metric, and learn the
 parameters of the value function that generates the risk metric and results in
 a policy that best matches the agent's observed
 behavior. The parameters of the value function, which ultimately drive the
 decision-making model,
are highly interpretable in terms of the degree of risk sensitivity
 and loss aversion. Thus, our technique supports prescriptive and
 descriptive analysis, both of which are important for the design of incentives
 and policies that takes into consideration the nuances of human decision-making behavior.

 Finally, as  noted in the introduction,  the results in this paper  significantly extend our previous
 work~\cite{mazumdar:2017aa}. We (i)  provide new convergence guarantees for
the  forward risk-sensitive reinforcement learning problem where behavioral-based value functions
 satisfy the
 assumptions and (ii)  provide more extensive theoretical results on the gradient-based
 inverse risk-sensitive
 reinforcement learning problem including proofs for theorems appearing 
 in the prior work. 

\section{Discussion}
\label{sec:discussion}
We present a new gradient based technique for learning risk-sensitive 
decision-making models of agents operating in uncertain environments. 
Specifically, we introduce a new forward risk-sensitive reinforcement learning
procedure with convergence guarantees for which  value functions retaining the fundamental shape of
behavioral value functions satisfy the assumptions. Based on this learning
algorithm, we introduce a well-formulated gradient-based inverse reinforcement learning
algorithm to recover the parameters indicating the observed agent's risk
preferences. 
We demonstrate the algorithm's
performance for agents based on several types of behavioral models and do so
on
two examples: the canonical Grid World problem and a passenger's view ride-sharing
where the parameters of the ride-sharing MDP are learned from real-world data.


Looking forward, we are examining ways of designing mechanisms to adaptively
produce the most
informative demonstrations and to incentivize certain behaviors. This is challenging since
some of the options in the state-action space may result in a very high risk to the
agent and thus, the volatility introduced by inducing such state-action pairs may be viewed unfavorably by the
agent causing them to completely opt-out.



%
\appendix
\subsection{Proof of Theorem~\ref{thm:contract}}
\label{app:thmContract}
The proof of Theorem~\ref{thm:contract} relies on the following fixed point theorem.
\begin{thm}[{Fixed Point Theorem~\cite[Theorem~2.2]{latif:2014aa}}]
    Let $(X,d)$ be a complete metric space and let $B_r(y)=\{x\in X|\
    d(x,y)<r\}$ be a ball of radius $r$, where $r>0$, centered at $y\in X$. Let $f:B_r(y)\rar X$ be a contraction
    map with contraction constant $h<1$. Further, assume that 
    $d(y,f(y))<r(1-h)$. Then, $f$ has a unqiue fixed point in $B_r(y)$.
    \label{thm:fix}
\end{thm}

\begin{IEEEproof}[Proof of Theorem~\ref{thm:contract}.a.]
    We claim that  $T$ is a contraction with constant
    $\bar{\alpha}=(1-\alpha(1-\gamma)\vep_K)$ where $\vep_K=\min\{D\tilde{v}(y)|
\ y\in I_K\}$.
    Indeed, 
    let $y(Q(x,a))=r(x,a,w)+\gamma \max_{a'}Q(x',a')-Q(x,a)$ be the temporal
    difference and define $g(x',a')=\max_{a'} Q(x',a')$. For any $Q\in B_K(0)$ we note that the temporal differences are bounded---in
    fact, $y(Q(x,a))\in I_K=[-M-K,M+K]$. Due to the monotonicity assumption on $v$, we have
    that for any $y',y\in I_K$, $\tilde{v}(y)-\tilde{v}(y')=\xi (y-y')$ for some
    $\xi\in [\vep_K,L]$. Recall the contraction map defined in
    \eqref{eq:Tupdate-1}:
    \begin{align}
        (TQ)(x,a)=&\alpha \mb{E}_{x',w}\big[
        \tilde{
        \uv}\big(y(Q(x,a))\big)\big]+Q(x,a)
        \label{eq:Tupdate-12}
    \end{align}
Then, for any $Q_1$ and $Q_2$, we have that
        \begin{align*}
            (TQ_1-TQ_2)(x,a)&=\alpha
            \mb{E}_{x',w}[\tilde{v}(y(Q_1(x,a))) -\tilde{v}(y(Q_2(x,a)))] +Q_1(x,a)-Q_2(x,a)\\
            &\leq
            \alpha\mb{E}_{x',w}[\xi_{x',w}(\gamma(g_1(x',a')+g_2(x',a'))-Q_1(x,a)+Q_2(x,a))]\\
            &\quad+Q_1(x,a)-Q_2(x,a)\\
            &\leq
            \alpha\gamma\mb{E}_{x',w}[\xi_{x',w}(g_1(x',a')+g_2(x',a'))]\\
            &\  +(1-\alpha \mb{E}_{x',w}[\xi_{x',w}])(Q_1(x,a)-Q_2(x,a)).
    \end{align*}
    Hence, 
    \begin{align*}
        |(TQ_1-TQ_2)(x,a)|&\leq
        (1-\alpha(1-\gamma)\mb{E}_{x',w}[\xi_{x',w}])\|Q_1-Q_2\|_\infty\\
        &\leq (1-\alpha(1-\gamma)\vep_K)\|Q_1-Q_2\|_\infty.
    \end{align*}
    We claim that the constant $\bar{\alpha}_K=1-\alpha(1-\gamma)\vep_K<1$. Indeed, recall that
    $0<\alpha\leq\min\{1,L^{-1}\}$ so that if
    $\alpha=L^{-1}$, then $\bar{\alpha}_K<1$ since $L=\max_{y\in I_K}D\tilde{v}(y)$ and
    $\vep_K=\min_{y\in I_K}D\tilde{v}(y)$. On the other hand, if $\alpha=1$, then
    $1\leq L^{-1}\leq (\vep_K)^{-1}$ so that $\vep_K\leq 1$ which, in turn,
    implies that 
    $\bar{\alpha}_K<1$. If $0<\alpha<\min\{1,L^{-1}\}$, then $\bar{\alpha}_K<1$
    follows trivially from the implications in the above two cases.
    Thus, $T$ is a contraction on $B_K(0)$ with the constant
    $\bar{\alpha}_K=(1-\alpha(1-\gamma)\vep_K)<1$. 
\end{IEEEproof}

\begin{IEEEproof}[Proof of Theorem~\ref{thm:contract}.b.]
Suppose 
    $K$ is chosen such that
    \begin{equation}\frac{\max\{|\tilde{v}(M)|,
        |\tilde{v}(-M)|\}}{1-\gamma}<K\min_{y\in
        I_K}D\tilde{v}(y).\label{eq:bound}\end{equation}
    Now, we argue that $T$ applied to the zero map, $0\in B_K(0)$, is strictly
    less than $K(1-\bar{\alpha}_K)$. Indeed, for any $\alpha\in (0, \min\{1,
    L^{-1}\}]$,
    \begin{align*}
        \|T(0)\|&\leq\alpha\max\{|v(M)|, |v(-M)|\}\\
        &<(1-\gamma)K\varepsilon_K\alpha=K(1-\bar{\alpha}_K)
    \end{align*}
    Combinging the above fact with the fact that $T$ is a contraction, the
    assumptions of Theorem~\ref{thm:fix} hold and, hence there is a unique fixed
    point $Q^\ast(x,a)\in B_K(0)$ for each $(x,a)\in X\times A$.
    \end{IEEEproof}

\subsection{Proof of Proposition~\ref{cor:valuefunctions}}
\label{app:cor}
Recall that $\tilde{v}\equiv v-v_0$. For different value functions,
Proposition~\ref{cor:valuefunctions} claims that the condition of
Theorem~\ref{thm:contract}.b,
\begin{equation}
        \frac{\max\{|\tilde{v}(M)|,
        |\tilde{v}(-M)|\}}{(1-\gamma)}<K\min_{y\in
        I_K}D\tilde{v}(y),\label{eq:boundass-lprospect}\end{equation}
holds.
\begin{IEEEproof}[{Proof of Proposition~\ref{cor:valuefunctions}.a}]
Suppose $v$ satisfies Assumption
\ref{ass:v} and that for some ${\varepsilon}>0$,            
${\varepsilon}<\frac{v(y)-v(y')}{y-y'}$ for all $y\neq y'$.
Then there exists a value of $K$, say $\bar K$, such
that~\eqref{eq:boundass-lprospect} holds for all $K>\bar K$. Indeed since
$\min_{K>0} \vep_K >{\varepsilon}$, for all $K$ satisfying
\[ \frac{\max\{|\tilde{v}(M)|,|\tilde{v}(-M)|\}}{{\varepsilon}(1-\gamma)}<K, \]
\eqref{eq:boundass-lprospect} must hold. 
\end{IEEEproof}
\begin{IEEEproof}[Proof of Proposition~\ref{cor:valuefunctions}.b]
We now show that for the {\lprospect } value function,
\eqref{eq:boundass-lprospect} holds for any choice of parameters
$(k_-,k_+,\zeta_-,\zeta_+)$. Indeed, for
$\zeta_+,\zeta_-\ge 1$ and any choice of $k_-,k_+$,
\[ \min_{K>0} \vep_K >{\varepsilon}>0 \]
where ${\varepsilon}=\min \{ \lim_{y\uparrow 0} D\tilde v(y), \lim_{y\downarrow 0} D\tilde v(y) \}$.
    Therefore, with $\zeta_+,\zeta_-\ge 1$, for any $K$ such that
    \[ \frac{\max\{|\tilde{v}(M)|,|\tilde{v}(-M)|\}}{{\vep}(1-\gamma)}<K, \]
\eqref{eq:boundass-lprospect} must hold. For the case when either $\zeta_+<1$ or
$\zeta_-<1$ or both, we note that
    \[ \min_{y \in I_K} D\tilde{v}(y)=\min\left\{\min_{y \in \{M+K,-M-K\}}
    D\tilde{v}(y), {\vep}\right\}.\]
    so that we need only show that for $\zeta_+<1$ and $\zeta_-<1$, there exists a $K$ such that 
    \begin{equation} \frac{\max\{|\tilde{v}(M)|,
        |\tilde{v}(-M)|\}}{1-\gamma}<KD\tilde{v}(K+M)
        \label{eq:plusbound}\end{equation}
    and
    \begin{equation} \frac{\max\{|\tilde{v}(M)|,
        |\tilde{v}(-M)|\}}{1-\gamma}<KD\tilde{v}(-K-M), \end{equation}
        respectively.
    Note that
    \[D\tilde{v}(y)=\left\{\begin{array}{ll}
            k_+\zeta_+(y-y_0+\epsilon)^{\zeta_+ -1}, & \ y\geq y_0\\
            k_-\zeta_-(y_0-y+\epsilon)^{\zeta_- -1}, & \
        y<y_0\end{array}\right.\] 
    Without loss of generality, we show \eqref{eq:plusbound} must hold for $\zeta_+<1$ and
    reference point $y_0=0$ (the proof for $\zeta_-<1$ follows an exactly
    analogous argument). 
    Plugging $D\tilde{v}(K+M)$ in and rearranging, we get that we need to find a $K$ such that 
    \[ \frac{\max\{|\tilde{v}(M)|,
        |\tilde{v}(-M)|\}}{(1-\gamma)\xi_+ k_+}<K (K+M+\epsilon)^{\xi_+-1}\] 
   Since the right-hand side above is a function of $K$ that is zero at $K=0$ and
   approaches infinity as $K \rar \infty$, and the left-hand side is a finite
   constant, there is some $\bar K$ such that for all $K>\bar K$, the above
   holds.    Thus, for the {\lprospect} value function, our assumptions are satisfied and
   there always exists a value of $K$ to choose in Theorem~\ref{thm:contract}.b.
\end{IEEEproof}

\begin{IEEEproof}[Proof of Proposition~\ref{cor:valuefunctions}.c]
    Suppose
    $v$ is an entropic map. 
    We note that, for the entropic map, $\min_{y\in I_K}D\tilde{v}(y)$ must
    occur at either $K+M$ or $-K-M$ if $\lambda<0$ or $\lambda>0$, respectively.
    Without loss of generality, let $\lambda>0$. 
    First, consider that the derivative of $\tilde{v}$,
    \[D\tilde{v}(y)=(\lambda)^{-1}e^{\lambda y},\]
    is minimized on $I_K$ at $-M-K$ for any $M$ and $K$. Moreover,
    $|\tilde{v}(M)|>|\tilde{v}(-M)|$. Hence, with
    $K=\lambda^{-1}$, we can derive conditions on $\lambda$ for which
    \eqref{eq:boundass-lprospect} holds. 
    In other words, with the specified $K$, we use \eqref{eq:boundass-lprospect} to
    dertermine which values of $\lambda$ are admissible. Indeed, from
    \eqref{eq:boundass-lprospect}, we have
    \[ (1-\gamma)^{-1} e^{\lambda M}<(\lambda)^{-1}e^{-\lambda M-1} \] 
   which 
   reduces to
 $ \lambda e^{2\lambda M}<(1-\gamma)e^{-1}.$
 Let $x=2\lambda M$, so that
  \[ x e^{x}<2M(1-\gamma)e^{-1}. \]
  Now, we can apply the Lambert $W$ function which satisfies $W(xe^x)=x$ for
  $x\geq 0$, to get that 
  \[ x<W(2M(1-\gamma)e^{-1}), \]
  so that
  \[ \lambda<(2M)^{-1}{W(2M(1-\gamma)e^{-1})}. \]
   Thus, if $|\lambda| < (2M)^{-1}{W(2M(1-\gamma)e^{-1})}$,
    then for the choice $K=\frac{1}{\lambda}$, \eqref{eq:boundass-lprospect}
    holds so that
    Theorem~\ref{thm:contract}.b holds for the entropic map.
\end{IEEEproof}

\subsection{Proof of Theorem~\ref{thm:DQ}}
\label{app:thmDQ}

We remark that the gradient algorithm and Theorem~\ref{thm:DQ} are
consistent with the gradient descent framework which uses the 
\emph{contravariant} gradient for learning as introduced in~\cite{amari:1998aa}
for Riemannian parameter spaces $\Theta$. Of course, when $\Theta$ is Euclidean
and the coordinate system is orthonormal, the gradient we normally use
(\emph{covariant} derivative) coincides with the contravariant gradient.
However, using the covariant derivative does not generalize to admissible parameter
spaces with more structure. 

Moreover, as is pointed out in~\cite{neu:2007aa},
the trajectories that result from the solution to the gradient algorithm are
equivalent up to reparameterization through a smooth invertible mapping with a
smooth inverse. Contravariant gradient methods have been shown to be
asymptotically efficient in a probabilistic sense and thus, they tend to avoid
\emph{plateaus}~\cite{amari:1998aa,peters:2005aa}.

Before we dive into the proof of Theorem~\ref{thm:DQ}, let us introduce some
definitions and useful propositions.

\begin{defn}[Fr{\'{e}}chet Subdifferentials] Let $U$ be a Banach space and $U^\ast$
    its dual. The Fr{\'{e}}chet subdifferential of $f: U\rar \mb{R}$ at $u\in
    U$, denoted by $\partial f(u)$ is the set of $u^\ast\in U^\ast$ such that
    \begin{equation}
        \lim_{h\rar 0}\inf_{h\neq 0} \|h\|^{-1}\left( f(u+h)-f(u)-\langle
        u^\ast, u\rangle \right)\geq 0.
        \label{eq:def}
    \end{equation}
\end{defn}
\begin{prop}[\cite{neu:2007aa,kruger:2003aa}]
    For a finite family $(f_i)_{i\in I}$ of real-valued functions 
    (where $I$ is a finite index set) defined on $U$, let $f(u)=\max_{i\in I}
    f_i(u)$. If $u^\ast\in\partial f_i(u)$ and $f_i(u)=f(u)$, then $u^\ast\in
    \partial f_i(u)$. If $f_1,f_2: U\rar \mb{R}$, $\alpha_1,\alpha_2\geq 0$,
    then $\alpha_1\partial f_1+\alpha_2\partial f_2 \subset \partial
    (\alpha_1f_1+\alpha_1f_2)$.
    \label{prop:subdiff}
\end{prop}
\begin{prop}[\cite{neu:2007aa,penot:1995aa}]
    Suppose that $(f_n)_{n\in \mb{N}}$ is a sequence of real-valued functions on
    $U$ which converge pointwise to $f$. Let $u\in U$, $u_n^\ast\in \partial
    f_n(u)\subset U^\ast$ and suppose that $(u_n^\ast)$ is weak$^\ast$--convergent to
    $u^\ast$ and is bounded. Moreover, suppose that at $u$, for any $\vep>0$,
    there exists an $N>0$ and $\delta>0$ such that for any $n\geq N$, $h\in
    B_U(0,\delta)$, a $\delta$--ball around
    $0$, $f_n(u+h)\geq f_n(u)+\langle u_n^\ast,h\rangle -\vep \|h\|$. Then
    $u^\ast\in \partial f(u)$.
    \label{prop:subdiffconverge}
\end{prop}

We now provide the proof for parts (a) and (b) of Theorem~\ref{thm:DQ}.
\begin{IEEEproof}[Proof of Theorem~\ref{thm:DQ}.a.]
    Let $Q_0(x,a,\theta)\equiv 0$. Then it is trivial that $Q_0(x,a,\theta)$ is
    locally Lipschitz in $\theta$ on $\Theta$. Supposing that $Q_t(x,a,\theta)$
    is $L_t$--locally Lipschitz in $\theta$, then we need to show that $TQ_t(x,a,\theta)$
    is locally Lipschitz which we recall is   defined by \begin{equation}
        (TQ)(x,a,\theta)=\alpha \mb{E}_{x',w}
        \tilde{
        \uv}(y(\theta),\theta)+Q(x,a, \theta)
        \label{eq:Tupdate}
    \end{equation}
    where $y(\theta)=r(x,a,w)+\gamma \max_{a'\in
    A}Q(x',a',\theta)-Q(x,a,\theta)$.

    Since $\tilde{\uv}\equiv \uv-\uv_0$, it also satisfies Assumption~\ref{ass:u}. Let
    $L_y=\max\{L_y(\theta)| \theta\in \Theta\}$ and define
    $g_t(x,\theta)=\max_{a'}Q_t(x,a',\theta)$. Note that since $Q_t$ is assumed
    Lipschitz with constant $L_t$, so is $g_t$. Surpressing the dependent of
    $TQ$ on $(x,a)$,  we have that
    \begin{align*}
        TQ_t(\theta)-TQ_t(\theta')&=\alpha \mb{E}_{x',w}[\tilde{
        \uv}(y(\theta),\theta)-\tilde{
        \uv}(y(\theta'),\theta')]+Q_t(x,a,\theta)-Q_t(x,a,\theta')\\
        &=\alpha\mb{E}_{x',w}[\tilde{v}(y(\theta),\theta)-\tilde{v}(y(\theta'),\theta)+\tilde{v}(y(\theta'),\theta)-\tilde{
        \uv}(y(\theta'),\theta')]\\
        &\qquad +Q_t(\theta)-Q_t(\theta').
    \end{align*}
    
Due to the monotonicity of $\tilde{v}$ in $y$, we know that for all
    $y_1,y_2$ there exists $\xi\in[\vep,L_y]$ such that
    \[\tilde{v}(y_1,\theta)-\tilde{v}(y_2,\theta)=\xi(y_1-y_2).\]
    Hence, 
\begin{align*}
    &    \mb{E}_{x',w}[\tilde{v}(y(\theta),\theta)-\tilde{v}(y(\theta'),\theta)+\tilde{v}(y(\theta'),\theta)-\tilde{
        \uv}(y(\theta'),\theta')]\\
        &\ \ = \mb{E}_{x',w}[\xi_{x',w}(y(\theta)-y(\theta'))+\tilde{v}(y(\theta'),\theta)-\tilde{
        \uv}(y(\theta'),\theta')]
    \end{align*}
    where we simply denote the dependence of $\xi$ on $x'$ and $w$, the components subject to
    randomness.
Then,
    \begin{align*} 
        TQ_t(\theta)-TQ_t(\theta')&=\alpha\mb{E}_{x',w}\big[\xi_{x',w}(y(\theta)-y(\theta'))+\tilde{v}(y(\theta'),\theta)-\tilde{
        \uv}(y(\theta'),\theta')\big]\\
       &\ \  \ \  +Q_t(\theta)-Q_t(\theta')\\
        &=\alpha\gamma\mb{E}_{x',w}[\xi_{x',w}(g_t(x',\theta)-g_t(x',\theta'))]-\alpha\mb{E}_{x',w}[\xi_{x',w}(Q_t(\theta)-Q_t(\theta'))]
    \\
&\ \ \ \ + \alpha \mb{E}_{x',w}[\tilde{v}(y(\theta'),\theta) -\tilde{
        \uv}(y(\theta'),\theta')] +Q_t(\theta)-Q_t(\theta')\\
        &\quad
        =\alpha\gamma\mb{E}_{x',w}[\xi_{x',w}(g_t(x',\theta)-g_t(x',\theta'))]-\alpha\mb{E}_{x',w}[\xi_{x',w}](Q_t(\theta)    -Q_t(\theta'))
\\
    &\ \ \ \ +\alpha \mb{E}_{x',w}[\tilde{v}(y(\theta'),\theta)-\tilde{
        \uv}(y(\theta'),\theta')] +Q_t(\theta)-Q_t(\theta')\\
        &\quad =(1-\alpha\mb{E}_{x',w}[\xi_{x',w}])(Q_t(\theta)
    -Q_t(\theta'))+\alpha\gamma\mb{E}_{x',w}[\xi_{x',w}(g_t(x',\theta)-g_t(x',\theta'))]\\
    &\ \ \ \ +\alpha \mb{E}_{x',w}[\tilde{v}(y(\theta'),\theta)-\tilde{\uv}(y(\theta'),\theta')]
    \end{align*}
    so that
    \begin{align*}
      \| TQ_t(\theta)-TQ_t(\theta')\| \leq &( (1-\alpha(1-\gamma) \vep)+\alpha
      L_\theta)L_t\|\theta-\theta'\|.
  \end{align*}
  Hence, letting $\bar{\alpha}=1-\alpha
    (1-\gamma)\vep$, we have that $TQ_t(\cdot, \cdot, \theta)$ is
    $L_{t+1}$--locally Lipschitz with $L_{t+1}=\bar{\alpha}L_t+\alpha L_\theta$.
    With $L_0=0$, by iterating, we get that
       \begin{align*}
           L_{t+1}&=(\bar{\alpha}^{t}+\cdots +\bar{\alpha}+1)\alpha L_\theta.
       \end{align*}
       As stated in Section~\ref{sec:irlgrad}, $T$ is a contraction so that
       $T^nQ_0\rar Q^\ast_\theta=Q^\ast(\cdot, \cdot,
       \theta)$ as $n\rar\infty$.
Hence, by the above argument, $Q^\ast_\theta$ is $\alpha
L_\theta/(1-\bar{\alpha})$--Lipschitz continuous.
\end{IEEEproof}
\begin{IEEEproof}[Proof of Theorem~\ref{thm:DQ}.b.]
   Consider a fixed vector $\theta\in \mb{R}^d$.  We now show that the
      operator $S$ acting on the space of functions $\phi_\theta:X\times A\rar
      \mb{R}^d$ and 
      defined 
by
\begin{align}
           (S\phi_\theta)(x,a)
           &=\alpha \mb{E}_{x',w}\big[ D_2
           \tilde{\uv}(y(\theta),\theta)+D_1\tilde{\uv}(y(\theta), \theta)(\gamma
       \phi_\theta(x',a_{x'}^\ast)-\phi_\theta(x,a)) \big]+\phi_\theta(x,a)
           \label{eq:Scontraction}
       \end{align}
              is a contraction where $a_{x'}^\ast$  is the action that maximizes $\sum_{a'\in
       A}\pi(a|x)Q(x,a,\theta)$ for any greedy policy $\pi$ with respect to
       $Q_\theta$.
              Indeed,
\begin{align*}
    (S\phi_\theta-S\phi_\theta')(x,a)&=\alpha
    \mb{E}_{x',w}[
        D_1\tilde{\uv}(y(\theta),\theta)\big(\gamma
        (\phi_\theta(x',a^\ast_{x'})-\phi_\theta'(x',a^\ast_{x'}))
   \notag\\
&\qquad -(\phi_\theta(x,a)-\phi_\theta'(x,a))\big)]+\phi_\theta(x,a)-\phi_\theta'(x,a)\\
&\leq (1-\alpha(1-\gamma)\mb{E}_{x',w}[D_1\tilde{\uv}(y(\theta),\theta)])\|\phi_\theta-\phi'_\theta\|_\infty
\end{align*}
so that, by Assumption~\ref{ass:u},
\[\|(S\phi_\theta- S\phi_\theta')(x,a)\|\leq
(1-\alpha(1-\gamma)\vep)\|\phi_\theta-\phi_\theta'\|_\infty.\]
Thus, $\bar{\alpha}$ is the required constant for ensuring $S$ is a contraction.
We remark that $S$ operates on each of the $d$ components of $\theta$ separately
and hence, it is a contraction when restricted to each individual component.

Let $\pi$ denote a greedy policy with respect to $Q^\ast_\theta$ and let $\pi_n$
be a sequence of policies that are greedy with respect to $Q_n=T^nQ_0$ where
ties are broken so that $\sum_{(x,a)\in X\times A}|\pi(a|x)-\pi_n(a|x)|$ is
minimized. Then for large enough $n$, $\pi_n=\pi$. Denote by $S_{\pi_n}$ the map
$S$ defined in \eqref{eq:Scontraction} where $\pi_n$ is the implemented policy. Consider the sequence
$\phi_{\theta,n}$ such that $\phi_{\theta,0}=0$ and
$\phi_{\theta,n+1}=S_{\pi_n}\phi_{\theta,n}$. For large enough $n$,
$\phi_{\theta, n+1}=S_{\pi}\phi_{\theta,n}$. 
Applying the (local) contraction mapping theorem (see,
\emph{e.g.},~\cite[Theorem~3.18]{sastry:1999aa}) we get that
$\lim_{n\rar\infty} S^n\phi_0$
converges to a unique fixed point.

Moreover, by induction and Proposition~\ref{prop:subdiff}, $\phi_{\theta,n}(x,a)\in
\partial_\theta Q_n(x,a,\theta)$. Hence, by
Proposition~\ref{prop:subdiffconverge}, the limit is a subdifferential of
$Q_\theta^\ast$ since $\tilde{\uv}$ is Lipschitz on $Y$ and $\Theta$ and the
derivatives of $\tilde{\uv}$ are uniformly bounded.
Since by part (a), $Q^\ast_\theta$ is locally Lipschitz in $\theta$, Rademacher's
Theorem (see, \emph{e.g.},~\cite[Thm.~3.1]{heinonen:2004aa}) tells us it is differentiable almost everywhere (expcept a set of Lebesgue measure zero). Since $Q_\theta^\ast$ is
differentiable, its subdifferential is its derivative.
\end{IEEEproof}

\subsection{Proof of Theorem~\ref{thm:DQball}}
\label{app:thmDbound}
Given that the proof  of Theorem~\ref{thm:DQball}
follows the same techniques as in Theorem~\ref{thm:contract} and
Theorem~\ref{thm:DQ}, we provide largely an outline, directing the reader to the
particular analogous components of the two proceeding theorems that are mimicked
in creating the proof.

\begin{IEEEproof}[Proof of Theorem~\ref{thm:DQball}.a.]
 For each $\theta$, the proof that $TQ(x,a,\theta)$
    is a contraction, and thus has a fixed point $Q^\ast_\theta\in B_K(0)$,
    follows directly that of Theorem~\ref{thm:contract} where instead of $Q_1$
    and $Q_2$ we have $Q(\theta)$ and $Q(\theta')$.  Given that $T$ is a
    contraction, the proof that $Q^\ast_\theta\in B_K(0)$ is Lipschitz with
    constant $\alpha L_\theta/(1-\bar{\alpha}_K)$ follows a
    similar argument to Theorem~\ref{thm:DQ}. 
\end{IEEEproof}
\begin{IEEEproof}[Proof of Theorem~\ref{thm:DQball}.b.]
The proof that $S$ is a contraction on $B_K(0)$ follows a similar argument to
that of Theorem~\ref{thm:DQ}, part (b). Indeed, \begin{align*}
    (S\phi_\theta-S\phi_\theta')(x,a) &=\alpha
    \mb{E}_{x',w}[
        D_1\tilde{\uv}(y(\theta),\theta)\big(\gamma
        (\phi_\theta(x',a^\ast_{x'})-\phi_\theta'(x',a^\ast_{x'}))
   \notag\\
&\qquad -(\phi_\theta(x,a)-\phi_\theta'(x,a))\big)]+\phi_\theta(x,a)-\phi_\theta'(x,a)\\
&\leq (1-\alpha(1-\gamma)\mb{E}_{x',w}[D_1\tilde{\uv}(y(\theta),\theta)])\|\phi_\theta-\phi'_\theta\|_\infty
\end{align*}
so that, by Assumption~\ref{ass:v},
\[\|(S\phi_\theta- S\phi_\theta')(x,a)\|\leq
(1-\alpha(1-\gamma)\vep_K)\|\phi_\theta-\phi_\theta'\|_\infty\] where
$\vep_K=\min\{D_1v(y,\theta)|\ y\in I_K\}$. Note that
$\bar{\alpha}_K=1-\alpha(1-\gamma)\vep_K<1$ for the same reasons as given in the
proof of Theorem~\ref{thm:contract} since $\alpha\in(0,\min\{1,L^{-1}\}]$.


For each $\theta\in \Theta$,            let $B_K(0)$ be the
            ball with radius $K$ satisfying
            \[\frac{\max\{|\tilde{v}(M,\theta)|,| \tilde{v}(-M,
            \theta)|\}}{1-\gamma}<K\min_{y\in I_K}
        D\tilde{v}(y,\theta).\]
            Then, for each $\theta$, $S$ satisfies Theorem~\ref{thm:fix} so that
            it has a unique fixed point in $B_K(0)$. 
            
            Following the same argument
            as in the proof of Theorem~\ref{thm:DQ}, part (b),
            by induction and Proposition~\ref{prop:subdiff}, $\phi_{\theta,n}(x,a)\in
\partial_\theta Q_n(x,a,\theta)$. Hence, by
Proposition~\ref{prop:subdiffconverge}, the limit is a subdifferential of
$Q_\theta^\ast$.
By part (a), $Q^\ast_\theta$ is locally Lipschitz in $\theta$ so that Rademacher's
Theorem (see, \emph{e.g.},~\cite[Thm.~3.1]{heinonen:2004aa}) implies it is differentiable almost everywhere (expcept a set of Lebesgue measure zero). Since $Q_\theta^\ast$ is
differentiable, its subdifferential is its derivative.

\end{IEEEproof}

\balance

\bibliographystyle{IEEEtran}
\bibliography{2017tacrefs}

\begin{thebibliography}{10}
\providecommand{\url}[1]{#1}
\csname url@samestyle\endcsname
\providecommand{\newblock}{\relax}
\providecommand{\bibinfo}[2]{#2}
\providecommand{\BIBentrySTDinterwordspacing}{\spaceskip=0pt\relax}
\providecommand{\BIBentryALTinterwordstretchfactor}{4}
\providecommand{\BIBentryALTinterwordspacing}{\spaceskip=\fontdimen2\font plus
\BIBentryALTinterwordstretchfactor\fontdimen3\font minus
  \fontdimen4\font\relax}
\providecommand{\BIBforeignlanguage}[2]{{%
\expandafter\ifx\csname l@#1\endcsname\relax
\typeout{** WARNING: IEEEtran.bst: No hyphenation pattern has been}%
\typeout{** loaded for the language `#1'. Using the pattern for}%
\typeout{** the default language instead.}%
\else
\language=\csname l@#1\endcsname
\fi
#2}}
\providecommand{\BIBdecl}{\relax}
\BIBdecl

\bibitem{koszegi:2006aa}
B.~K{\"o}szegi and M.~Rabin, ``A model of reference-dependent preferences,''
  \emph{The Quarterly J.~Economics}, vol. 121, no.~4, pp. 1133--1165, 2006.

\bibitem{tversky:1991aa}
A.~Tversky and D.~Kahneman, ``Loss aversion in riskless choice: A
  reference-dependent model,'' \emph{The Quarterly J.~Economics}, vol. 106,
  no.~4, pp. 1039--1061, 1991.

\bibitem{tversky:1986aa}
------, ``Rational choice and the framing of decisions,'' \emph{J.~Business},
  vol.~59, no.~4, pp. pp. S251--S278, 1986.

\bibitem{geibel:2005aa}
P.~Geibel and F.~Wysotzki, ``Risk-sensitive reinforcement learning applied to
  control under constraints,'' \emph{J.~Artificial Intelligence Research},
  vol.~24, pp. 81--108, 2005.

\bibitem{borkar:2002ab}
V.~S. Borkar and S.~P. Meyn, ``Risk-sensitive optimal control for markov
  decision processes with monotone cost,'' \emph{Mathematics of Operations
  Research}, vol.~27, no.~1, pp. 192--209, 2002.

\bibitem{l.a.:2016aa}
P.~L.A., C.~Jie, M.~Fu, S.~Marcus, and C.~Szepesv{\'{a}}ri, ``Cumulative
  prospect theory meets reinforcement learning: Prediction and control,'' in
  \emph{Proc. 33rd Intern.~Conf.~on Machine Learning}, vol.~48, 2016.

\bibitem{shen:2014aa}
Y.~Shen, M.~J. Tobia, and K.~Obermayer, ``Risk-sensitive reinforcement
  learning,'' \emph{Neural Computation}, vol.~26, pp. 1298--1328, 2014.

\bibitem{mihatsch:2002aa}
O.~Mihatsch and R.~Neuneier, ``Risk-sensitive reinforcement learning,''
  \emph{Machine Learning}, vol.~49, no.~2, pp. 267--290, 2002.

\bibitem{nagengast:2010aa}
A.~J. Nagengast, D.~A. Braun, and D.~M. Wolpert, ``Risk-sensitive optimal
  feedback control accounts for sensorimotor behavior under uncertainty,''
  \emph{PLOS Computational Biology}, vol.~6, no.~7, pp. 1--15, 2010.

\bibitem{majumdar:2017aa}
A.~Majumdar, S.~Singh, A.~Mandlekar, and M.~Provone, ``Risk-sensitive inverse
  reinforcement learning via coherent risk models,'' in \emph{{Robotics:
  Science and Systems}}, 2017.

\bibitem{ng:2000aa}
A.~Y. Ng and S.~Russell, ``{Algorithms for Inverse Reinforcement Learning},''
  in \emph{Proc.~17th Inter.~Conf.~Machine Learning}, 2000, pp. 663--670.

\bibitem{abbeel:2004aa}
P.~Abbeel and A.~Y. Ng, ``Apprenticeship learning via inverse reinforcement
  learning,'' in \emph{Proc.~21st Inter.~Conf.~Machine Learning}, 2004.

\bibitem{ratliff:2006aa}
N.~D. Ratliff, J.~A. Bagnell, and M.~A. Zinkevich, ``Maximum margin planning,''
  in \emph{Proc.~23rd Inter.~Conf.~Machine Learning}, 2006, pp. 729--736.

\bibitem{shen:2013aa}
Y.~Shen, W.~Stannat, and K.~Obermayer, ``{Risk-Sensitive Markov Control
  Processes},'' \emph{SIAM J. Control Optimization}, vol.~51, no.~5, pp.
  3652--3672, 2013.

\bibitem{mazumdar:2017aa}
E.~Mazumdar, L.~J. Ratliff, T.~Fiez, and S.~S. Sastry, ``Gradient-based inverse
  risk-sensitive reinforcement learning,'' in \emph{Proc.~56th IEEE
  Conf.~Decision and Control}, 2017.

\bibitem{heger:1994aa}
M.~Heger, ``Consideration of risk in reinforcement learning,'' in
  \emph{Proc.~11th Inter.~Conf.~Machine Learning}, 1994, pp. 105--111.

\bibitem{kahneman:1979aa}
D.~Kahneman and A.~Tversky, ``Prospect theory: An analysis of decision under
  risk,'' \emph{Econometrica}, vol.~47, no.~2, pp. 263--291, 1979.

\bibitem{tversky:1992aa}
A.~Tversky and D.~Kahneman, ``Advances in prospect theory: Cumulative
  representation of uncertainty,'' \emph{J.~Risk and Uncertainty}, vol.~5,
  no.~4, pp. 297--323, Oct 1992.

\bibitem{gonzalez:1999aa}
R.~Gonzalez and G.~Wu, ``On the shape of the probability weighting function,''
  \emph{Cognitive Psychology}, vol.~38, no.~1, pp. 129--166, 1999.

\bibitem{simon:2000aa}
H.~Simon, ``{Bounded rationality in social science: Today and tomorrow},''
  \emph{Mind \& Society}, vol.~1, no.~1, pp. 25--39, Mar. 2000.

\bibitem{tversky:1981aa}
A.~Tversky and D.~Kahneman, ``{The framing of decisions and the psychology of
  choice},'' \emph{Science}, vol. 211, no. 4481, pp. 453--458, Jan. 1981.

\bibitem{camerer:1989aa}
C.~F. Camerer, ``An experimental test of several generalized utility
  theories,'' \emph{J.~Risk and Uncertainty}, vol.~2, no.~1, pp. 61--104, 1989.

\bibitem{follmer:2002aa}
H.~F{\"o}llmer and A.~Schied, ``Convex measures of risk and trading
  constraints,'' \emph{Finance and Stochastics}, vol.~6, no.~4, pp. 429--447,
  2002.

\bibitem{coraluppi:2000aa}
S.~P. Coraluppi and S.~I. Marcus, ``Mixed risk-neutral/minimax control of
  discrete-time, finite-state markov decision processes,'' \emph{IEEE
  Trans.~Autom.~Control}, vol.~45, no.~3, pp. 528--532, 2000.

\bibitem{artzner:1999aa}
P.~Artzner, F.~Delbaen, J.-M. Eber, and D.~Heath, ``Coherent measures of
  risk,'' \emph{Mathematical Finance}, vol.~9, no.~3, pp. 203--228, 1999.

\bibitem{robbins:1985aa}
H.~Robbins and D.~Siegmund, \emph{A Convergence Theorem for Non Negative Almost
  Supermartingales and Some Applications}.\hskip 1em plus 0.5em minus
  0.4em\relax Springer New York, 1985, pp. 111--135.

\bibitem{robbins:1951aa}
H.~Robbins and S.~Monro, ``A stochastic approximation method,'' \emph{The
  Annals of Mathematical Statistics}, vol.~22, no.~3, pp. 400--407, 1951.

\bibitem{tsitsiklis:1994aa}
J.~N. Tsitsiklis, ``Asynchronous stochastic approximation and q-learning,''
  \emph{Machine Learning}, vol.~16, no.~3, pp. 185--202, 1994.

\bibitem{kushner:2003aa}
H.~J. Kushner and G.~G. Yin, \emph{Stochastic Approximation and Recursive
  Algorithms and Applications}.\hskip 1em plus 0.5em minus 0.4em\relax
  Springer, 2003.

\bibitem{neu:2007aa}
G.~Neu and C.~Szepesv\'{a}ri, ``Apprenticeship learning using inverse
  reinforcement learning and gradient methods,'' in \emph{Proc.~23rd
  Conf.~Uncertainty in Artificial Intelligence}, 2007, pp. 295--302.

\bibitem{heinonen:2004aa}
J.~Heinonen, ``{Lectures on Lipschitz Analysis},'' \emph{14th
  Jyv{\"{a}}skyl{\"{a}} Summer School}, 2004.

\bibitem{millar:1979aa}
P.~W. Millar, ``Asymptotic minimax theorems for the sample distribution
  function,'' \emph{Zeitschrift f{\"{u}}r Wahrscheinlichkeitstheorie und
  Verwandte Gebiete}, vol.~48, no.~3, pp. 233--252, 1979.

\bibitem{massart:1990aa}
P.~Massart, ``The tight constant in the dvoretzky-kiefer-wolfowitz
  inequality,'' \emph{The Annals of Probability}, vol.~18, pp. 1269--1283,
  1990.

\bibitem{latif:2014aa}
A.~Latif, \emph{Banach Contraction Principle and Its Generalizations}.\hskip
  1em plus 0.5em minus 0.4em\relax Springer International Publishing, 2014, pp.
  33--64.

\bibitem{amari:1998aa}
S.-I. Amari, ``Natural gradient works efficiently in learning,'' \emph{Neural
  Computation}, vol.~10, no.~2, pp. 251--276, 1998.

\bibitem{peters:2005aa}
J.~Peters, S.~Vijayakumar, and S.~Schaal, ``Natural actor-critic,'' in
  \emph{Proc.~16th European Conf.~Machine Learning}, 2005, pp. 280--291.

\bibitem{kruger:2003aa}
A.~Y. Kruger, ``{On Fr{\'{e}}chet Subdifferentials},'' \emph{J.~Mathematical
  Sciences}, vol. 116, no.~3, 2003.

\bibitem{penot:1995aa}
J.~Penot, ``On the interchange of subdifferentiation and epi-convergence,''
  \emph{J.~Mathematical Analysis and Applications}, vol. 196, no.~2, pp.
  676--698, 1995.

\bibitem{sastry:1999aa}
S.~S. Sastry, \emph{Nonlinear Systems}.\hskip 1em plus 0.5em minus 0.4em\relax
  Springer, 1999.

\end{thebibliography}




\end{document}